\documentclass[10pt,journal,compsoc]{IEEEtran}
\ifCLASSOPTIONcompsoc
  \usepackage[nocompress]{cite}
\else
  \usepackage{cite}
\fi
\ifCLASSINFOpdf
   \usepackage[pdftex]{graphicx}
\else
\fi
\usepackage{amsmath}
\usepackage{amsthm}
\usepackage{amsfonts}
\newtheorem{definition}{Definition}

\newtheorem{proposition}{Proposition}
\newtheorem{theorem}{Theorem}
\DeclareMathOperator*{\argmax}{arg\,max}
\usepackage{multicol}
\usepackage{tablefootnote}
\usepackage{booktabs}
\usepackage[inline]{enumitem}
\usepackage{url}
\usepackage{chngcntr}
\begin{document}
%
\title{Geometry- and Accuracy-Preserving Random Forest Proximities}

\author{Jake~S.~Rhodes, Adele~Cutler, and Kevin~R.~Moon
\IEEEcompsocitemizethanks{
\IEEEcompsocthanksitem Jake~S.~Rhodes is with Idaho State University
\IEEEcompsocthanksitem Adele~Cutler and Kevin~R.~Moon are with Utah State University
\IEEEcompsocthanksitem Email: jakerhodes@isu.edu, adele.culter@usu.edu, kevin.moon@usu.edu}
\thanks{}}

\IEEEtitleabstractindextext{%
\begin{abstract}
Random forests are considered one of the best out-of-the-box classification and regression algorithms due to their high level of predictive performance with relatively little tuning. Pairwise proximities can be computed from a trained random forest and measure the similarity between data points relative to the supervised task. Random forest proximities have been used in many applications including the identification of variable importance, data imputation, outlier detection, and data visualization. However, existing definitions of random forest proximities do not accurately reflect the data geometry learned by the random forest. In this paper, we introduce a novel definition of random forest proximities called Random Forest-Geometry- and Accuracy-Preserving proximities (RF-GAP). We prove that the proximity-weighted sum (regression) or majority vote (classification) using RF-GAP exactly matches the out-of-bag random forest prediction, thus capturing the data geometry learned by the random forest. We empirically show that this improved geometric representation outperforms traditional random forest proximities in tasks such as data imputation and provides outlier detection and visualization results consistent with the learned data geometry.
\end{abstract}

}

\maketitle

\IEEEdisplaynontitleabstractindextext

%
\IEEEpeerreviewmaketitle

\IEEEraisesectionheading{\section{Introduction}\label{sec:introduction}}

%
%
%
%
\IEEEPARstart{R}ANDOM forests~\cite{Breiman2001randomforests} are well-known, powerful predictors comprised of an ensemble of binary recursive decision trees. Random forests are easily adapted for both classification and regression, are trivially parallelizable, can handle mixed variable types (continuous and categorical), are unaffected by monotonic transformations, are insensitive to outliers, scale to small and large datasets, handle missing values, are capable of modeling non-linear interactions, and are robust to noise variables~\cite{Breiman2001randomforests, Cutler2012randomforests}. Random forests are simple to use, produce good results with little to no tuning, and can be applied to a wide variety of fields. Citing some recent examples, Benali et al.~\cite{benali2019solar} demonstrated superior solar radiation prediction using random forests over neural networks. The work of~\cite{Han2018ComparisonOR} showed that random forests produced the best results with the lowest standard errors in classifying error types in rotating machinery when compared with more commonly used models in this application, such as the SVM and neural networks. Other recent successes include patient health prediction after exposure to COVID-19~\cite{iwendi2020covid}, spatio-temporal COVID-19 case estimation~\cite{YESILKANAT2020covidSpac}, landslide susceptibility mapping~\cite{nhu2020landslide}, infectious diarrhea forecasting~\cite{Fang2020diar}, cardiovascular disease prediction~\cite{Yang2020cardio}, deforestation rate prediction~\cite{saha2020deforest}, nanofluid viscosity estimation~\cite{GHOLIZADEH2020nanofluid}, rural credit assessment~\cite{rao2020credit}, RNA pseudouridine site prediction~\cite{Lv2020rna}, wearable-sensor activity classification~\cite{tahir2020sensor}, heavy metal distribution estimation in agricultural soils~\cite{tan2020soil}, and cell type classification via Raman spectroscopy~\cite{ZHANG2020221labelfree}.

In addition to their high predictive power, random forests have a natural extension to produce pair-wise proximity (similarity) measures determined by the partitioning space of the decision trees which comprise them. The random forest proximity measure between two observations was first defined by Leo Breiman as the proportion of trees in which the observations reside in the same terminal node~\cite{BreimanRandomForest:Online}. As splitting variable values are found to optimize partitioning according to the supervised task, these proximities encode a supervised similarity measure. 

Many machine learning methods depend on some measure of pairwise similarity (which is usually unsupervised) including dimensionality reduction methods~\cite{KruskalWish1978mds, vanDerMaaten2008tsne, Moon2019phate, duque2019dig, duque2020grae, Tenenbaum2000isomap, lel2018umap}, spectral clustering~\cite{ng2001spectral}, and any method involving the kernel trick such as SVM~\cite{Cortes1995svm} and kernel PCA~\cite{Scholkopf99kernelprincipal}. Random forest proximities can be used to extend many unsupervised problems to a supervised setting and have been used for data visualization~\cite{pouyan2016distancelearning, gray2011dementia, Shi2005tumor, pang2006pathways, rhodes2021rfphate}, outlier detection~\cite{pang2006pathways, zhang2008Network, gislason2006landcover, narayana2020outlierdet}, and data imputation~\cite{Kokla2019metabolomicsimputation, Ramosaj2019predictmiss, pantanowitz2009missingdata, Shah2014caliber}. See Section~\ref{sec:applications} for further discussion of random forest proximity applications. 

Unlike unsupervised similarity measures, random forest proximities incorporate variable importance relative to the supervised task as these variables are more likely to be used in determining splits in the decision trees~\cite{Cutler2012randomforests}. Ideally, random forest proximities should define a data geometry that is consistent with the learned random forest; that is, the random forest predictive ability should be recoverable from the proximities. In this case, applications involving random forest proximities, such as data visualization, can lead to improved interpretability of the random forests specifically, and more generally the data geometry relative to the supervised task. 

One way to test if the proximity-defined geometry is consistent with the learned random forest is to compute a proximity-weighted predictor where a data point's predicted label consists of a proximity-weighted sum of the labels of all other points. This predictor should match the random forest's predictions if the proximities are capturing the random forest's learning. In particular, matching the random forest's out-of-bag predictions is a good indication that the proximities capture the random forest's learned data geometry in an unbiased fashion as the out-of-bag samples were not used to construct the tree. Under Breiman's original definition, the proximity-weighed predictions do not match the random forest's out-of-bag predictions when applied to the training data (see Section \ref{sec:results}).  Thus, this definition does not capture the data geometry learned by the random forest, limiting its potential for improved interpretability of the random forest. Indeed, this inconsistency negatively affects random forest proximity applications such as data visualization and imputation. 

We define a new random forest proximity measure called Random Forest-Geometry- and Accuracy-Preserving proximities (RF-GAP) that defines a data geometry such that the proximity-weighted predictions exactly match those of the random forest for both regression and classification problems. Under our definition, an out-of-bag observation's proximities are computed via in-bag (training) observations. That is, the sample used to generate a decision tree also generates the proximities of out-of-bag observations. 

RF-GAP is purposefully designed to match the random forest predictions and follows the same schema for a classification or regression learning problem. Random forests use in-bag observations or  ``voting points''~\cite{lin2006adaptiveNN} within terminal nodes to make predictions. For our definition, the proximities between out-of-bag observations are weights generated using the counts of in-bag observations sharing the same terminal node. We prove the equivalence between the proximity-weighted predictions with those of the random forest and demonstrate this empirically. In this sense, RF-GAP proximities preserve the data geometry as viewed by the random forest's learning. We compare RF-GAP proximities with existing random forest proximities in common applications: data imputation, visualization, and outlier detection. In these applications, RF-GAP outperforms existing definitions, empirically showing clear benefits from the improved random forest geometry representation.

\section{Random Forest Proximity Applications}\label{sec:applications}

Random forest proximities have been used in many applications such as visualization via dimensionality reduction or clustering. Unsupervised random-forest clustering was introduced by Shi and Horvath in~\cite{horvath2006unsupervised}.  In this paper, they applied both classical and metric multi-dimensional scaling (MDS) to a number of datasets. Pouyan et al. used unsupervised random forest proximity matrices to generate 2D plots for visualization using t-SNE~\cite{pouyan2016distancelearning}. This approach was used to visualize two Mass cytometry data sets and provided clearer visual results over other compared distance metrics (Euclidean, Chebyshev, and Canberra). Similarly, \cite{Shi2005tumor} used an unsupervised random forest for clustering tumor profilings and showed improvement over other common clustering algorithms with microarray data. Random forest proximities have also been used in the supervised setting for visualizing the data via dimensionality reduction, typically via MDS~\cite{finehout2007cerebrospinal,pang2006pathways}, but other approaches have been used~\cite{rhodes2021rfphate}. However, these approaches use a proximity definition that does not match the data geometry learned by the random forest. Thus the visualizations do not give an accurate representation of the random forest's learning. 

Random forest proximities have also been used in outlier detection in a supervised setting. An observation's outlier score is typically defined to be inversely proportional to its average within-class proximity measure (see Section~\ref{subsec:outliers} for details). This approach was used to achieve better random forest error rates in both classification and regression in pathway analysis~\cite{pang2006pathways}. Nesa et al. demonstrated the superiority of random forests for detecting errors and events across the internet of things (IoT) device sensors over other multivariate outlier detection methods~\cite{nesa2018iot}. The random forest outlier detection algorithm has been effective in other contexts such as modeling species distribution~\cite{liu2018species}, detecting food adulteration via infrared spectroscopy~\cite{santana2019foodadult}, predicting galaxy spectral measurements~\cite{baron2016galaxy}, and detecting network anomalies~\cite{vejendla2020network}. 

Another application of random forest proximities is data imputation. This is accomplished by replacing missing values of a given variable with a proximity-weighted sum of non-missing values. Pantanowitz and Marwala used this approach to impute missing data in the context of HIV seroprevalence~\cite{pantanowitz2009missingdata}. They compared their results with five additional imputation methods, including neural networks, and random forest-neural network hybrids, and concluded that random forest imputation produced the most accurate results with the lowest standard errors. Shah et al. compared random forest imputation with multivariate imputation by chained equations (MICE~\cite{vanburren2011mice}) on cardiovascular health records data~\cite{Shah2014caliber}.  They showed that random forest imputation methods typically produced more accurate results and that in some circumstances MICE gave biased results under default parameterization. In~\cite{Kokla2019metabolomicsimputation} it was similarly shown that random forests produced the most accurate imputation in a comprehensive metabolomics imputation study.

Variable importance assessment is another application of random forest proximities. One approach is to use differences in proximity measures as a criterion for assessing variable permutation importance \cite{Zhou2010GeneSU}. Variable selection criteria were compared across various data sets and some improvement over existing variable importance measures was achieved. In \cite{whitmore2018explicating}, feature contributions to the random forest decision space (defined by proximities) are explored. While many measures of variable importance are generally computed at a global level, the authors propose a bitwise, permutation-based feature importance which captures both the contribution (influence of the feature in the decision space) and closeness (position in the decision space relative to the in- and out-class), giving further insight to the contribution of each feature at the terminal node level.

Multi-modal/multiview problems can also be approached using random forest proximities. Gray et al. additively combined random forest proximities from different modalities or views (FDG-PET and MR imaging) of persons with Alzheimer's disease or with mild cognitive impairment. They applied MDS to the combined proximities to create an embedding used for classification.  Classification on the multi-modal embedding showed significantly better results than classification on both modes separately~\cite{gray2011dementia}. Cao, Bernard, Sabourin, and Heutte explored variations of proximity-based classification techniques in the context of multi-view radiomics problems~\cite{cao2019multi-viewradiomics}. They compared with the work from~\cite{gray2011dementia} and joined proximity matrices using linear combinations. The authors explored random forest parameters to determine the quality of the proximity matrices. They concluded that a large number of maximum-depth trees produced the best quality proximities, quantified using a one-nearest neighbor classifier. Using our proposed random forest proximity measure which accurately reflects the random forest predictions from each view may add to the success of this method, thus creating a truer forest ensemble for multi-view learning.

Seoane et al. proposed the use of random forest proximities to measure gene annotations with some improvement in precision over other existing methods~\cite{seoane2014geneannotation}. Zhao et al. presented two matching methods for observational studies, one propensity-based and the other random forest-proximity-based~\cite{Zhao2016propensity}. For the random forest approach, subjects of different classes were iteratively matched according to the nearest proximity values. They showed that proximity-based matching was superior to propensity matching in addition to other existing matching techniques. 

All of these applications used existing definitions of random forest proximities that do not match the data geometry learned by the random forest. In contrast, RF-GAP accurately reflects this geometry. Thus, the applications presented here should show improvement using our definition. Indeed, we show experimentally in Section~\ref{sec:results} that using RG-GAP gives data visualizations that more accurately represent the geometry learned from the random forests, outlier scores that are more reflective of the random forest's learning, and improved random forest imputations. 


\section{Random Forest Proximities}\label{sec:proximities}

Here we provide several existing definitions of random forest proximities followed by RF-GAP which preserves the geometry learned by the random forest. Let $\mathcal{M} = \left\{ (\mathbf{x}_1, y_1), (\mathbf{x}_2, y_2), \cdots (\mathbf{x}_N, y_N) \right\}$ be the training data where each $\mathbf{x}_i \in \mathcal{X}$, is a $d$-dimensional vector of predictor variables with corresponding response, $y_i \in \mathcal{Y}$. 

The strength of the random forest is highly dependent on the predictive power of the individual decision trees (base learners) and on the low correlation between the decision trees~\cite{hastie2017elements,BreimanRandomForest:Online}. To decrease the correlation between the decision trees, a bootstrap sample of the training data is used to train each decision tree $t$. Each decision tree $t$ in a random forest is grown by recursively partitioning (splitting) the bootstrap sample into nodes, where splits are determined across a subset of feature variables to maximize purity (classification) or minimize the mean squares of the residuals (regression) in the resulting nodes. This process repeats until a stopping criterion is met. For classification, splits are typically continued until nodes are pure (one class). For regression, a common stopping criterion is a predetermined minimum node size (e.g., 5). The trees in random forests are typically not pruned.

In addition to bootstrap sampling, further correlational decrease between trees is ensured by selecting a random subset of predictor variables at each node for split optimization. The number of variables to be considered is designated by the parameter \texttt{mtry} in some random forest packages (e.g., \texttt{randomForest}~\cite{randomForest2021} and \texttt{ranger}~\cite{wright2017ranger}). 

\begin{figure}[htb!]
    \centering
    \includegraphics[width = .48\textwidth]{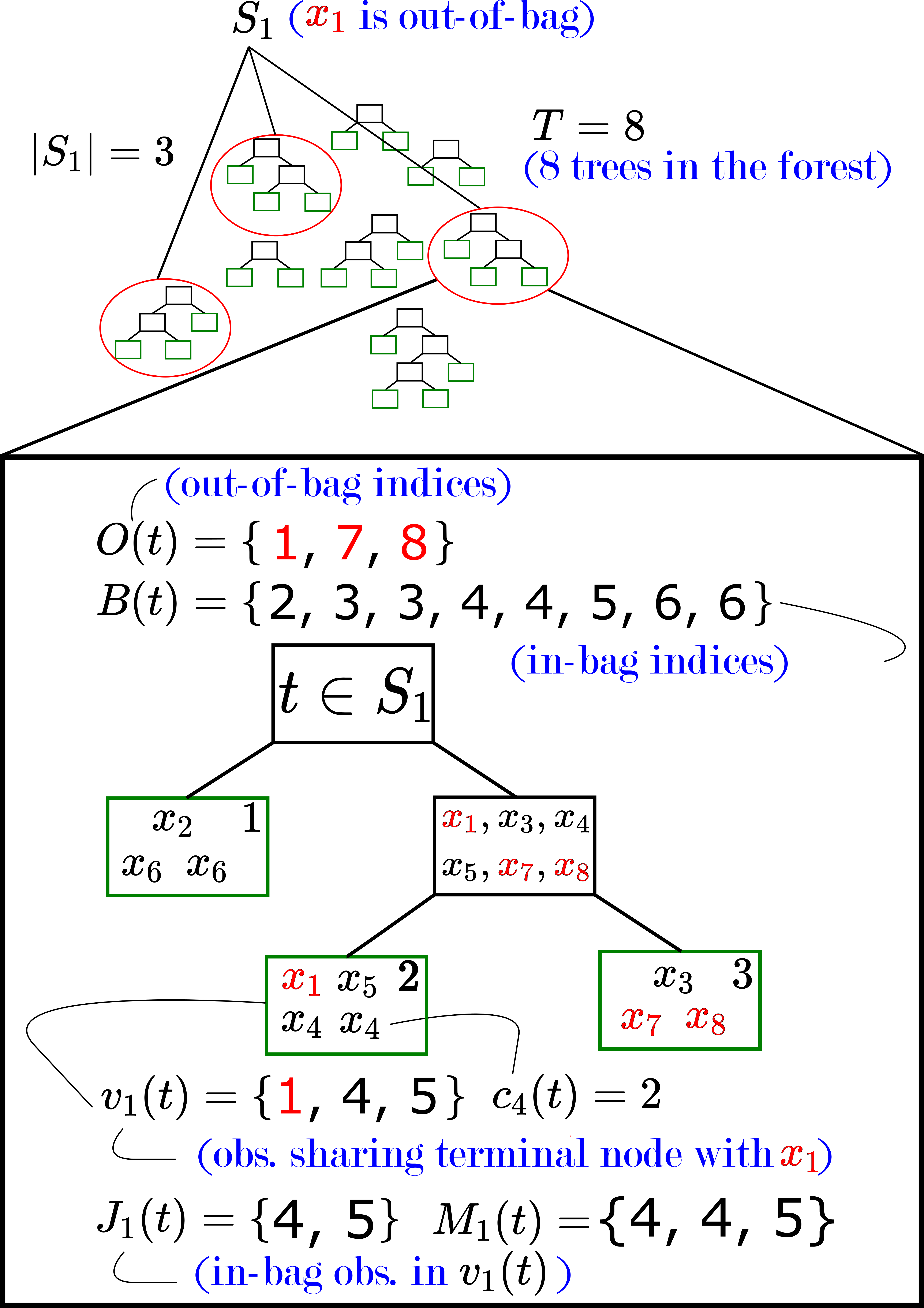}
    \caption{An example of a random forest and notation with regards to a particular observation $x_1$. The red-encircled trees are those in which $x_1$ is out of bag, making up the set of trees $S_1$. A particular tree in $S_1$ is exhibited. The out-of-bag indices for the tree are given in red $(i \in O(t))$, while the in-bag indices $(i \in B(t))$ are shown in black. The indices of observations residing in the same terminal node as $x_1$ are given by the set $v_1(t)$.  $J_1(t)$ gives the in-bag observation indices in the terminal node $v_1(t)$, while $M_1(t)$ provides the corresponding multiset. $c_4(t)$ denotes the number of repetitions of observation $x_4$ in the tree $t$.}
    \label{fig:notation}
\end{figure}

We will use the following notation to define random forest proximities (see Figure~\ref{fig:notation} for a visual example):

\begin{itemize}
    \item $\mathcal{T}$ is the set of decision trees in a random forest trained on $\mathcal{M}$ with $|\mathcal{T}| = T$.
    \item $B(t)$ is the multiset of indices in the bootstrap sample of the training data that is randomly selected to train the tree $t\in\mathcal{T}$. Thus $B(t)$ contains the indices of the in-bag observations.
    \item $O(t)=\{i=1,\dots,n|i\notin B(t) \}$. Thus $O(t)$ is the set of indices of the training data that are not contained in $B(t)$. $O(t)$ is often referred to as the out-of-bag (OOB) sample.
    \item $S_i=\{t\in\mathcal{T}|i\in O(t)\}$. This is the set of trees in which the $i$th observation  is OOB.
    \item $v_i(t)$ contains the indices of all observations that end up in the same terminal node as $\mathbf{x}_i$ in tree $t$.
    \item $J_i(t)=v_i(t)\cap B(t)$. This is the set of indices in $v_i(t)$ that correspond with the in-bag observations of $t$. I.e., these are the observations that are in-bag and end up in the same terminal node as $\mathbf{x}_i$.
    \item $M_i(t)$ is the multiset of in-bag indices in the terminal node shared with the $i$th observation in tree $t$ including multiplicities.
    \item $c_j(t)$ is the in-bag multiplicity of the observation $j$ in tree $t$. $c_j(t) := 0$ if observation $j$ is OOB. Thus, $\sum_{j = 1}^N c_j(t) = N$.
\end{itemize}


The terminal nodes of a random forest partition the input space $\mathcal{X}$. This partition is often used in defining random forest proximities as in Breiman's original definition:

\begin{definition}[Original Random Forest Proximity~\cite{BreimanRandomForest:Online}]
\label{def:training_prox}
The random forest proximity between observations $i$ and $j$ is given by:
\[p_{Or}(i, j) = \dfrac{1}{T}\sum_{t = 1}^T I(j\in v_i(t)),\]
where $T$ is the number of trees in the forest, $v_i(t)$ contains the indices  of observations that end up in the same terminal node as $\mathbf{x}_i$ in tree $t$, and $I(.)$ is the indicator function. That is, the proximity between observations $i$ and $j$ is the proportion of trees in which they reside in the same terminal node, regardless of bootstrap status.
\end{definition}

Definition \ref{def:training_prox} (the original definition) does not capture the data geometry learned by the random forest as it does not take an observation's bootstrap status (whether or not the observation was used in the training of any particular tree) into account in the proximity calculation: both in-bag and out-of-bag samples are used and equally weighted. In-bag observations of different classes will necessarily terminate in different nodes (as trees are grown until pure). Thus this produces an over-exaggerated class separation in the proximity values.
 


Despite these weaknesses, this definition has been used for outlier detection, data imputation, and visualization. However, these applications may produce misleading results as this definition tends to overfit the training data, quantified by low error rates as proximity-weighted predictors. One attempt to overcome this issue redefines the proximity measure between observations $i$ and $j$ using only trees in which both observations are out-of-bag (OOB proximities): 

\begin{definition}[OOB Proximity~\cite{Hastie2009RandomForest, randomForest2021}]
\label{def:oob_prox}
The OOB proximity between observations with indices $i$ and $j$ is given by:

\[p_{OOB}(i, j) = \dfrac{ \sum_{t\in S_i}I(j\in O(t)\cap v_i(t))}{ \sum_{t\in S_i}I(j\in O(t))},\]
where $O(t)$ denotes the set of indices of the observations that are out-of-bag in tree $t$, and $S_i$ is the set of trees for which observation $i$ is out-of-bag. In other words, this proximity measures the proportion of trees in which observations $i$ and $j$ reside in the same terminal node, both being out-of-bag. 
\end{definition}

Definition \ref{def:oob_prox}~\cite{Hastie2009RandomForest} is currently used in the \texttt{randomForest}~\cite{randomForest2021} package by Liaw and Wiener in the \texttt{R} programming language~\cite{R2021}. It may have been inadvertently used in papers that used this package but, none made explicit mention of the use of OOB observations in building proximities. However, we find that this definition also does not characterize the random forest predictions and generally produces higher error rates as a proximity-weighted predictor when compared to the random forest's OOB error rate (see Section~\ref{sec:results}). The reason that the results do not match is two-fold. First, this definition does not directly use in-bag observations. A classification or regression forest trains on a training set (in-bag set) and predicts on an unseen test set (OOB observations). Thus the OOB proximities are missing a key factor in the random forest behavior. Second, random forest predictions are made by voting or averaging in-bag labels. In contrast, the OOB proximities are generated without regard to the number of observations within a given terminal node.

Several alternative random forest proximity measures beyond Definitions~\ref{def:training_prox} and~\ref{def:oob_prox} have been proposed. In each case, source code has not been provided. In~\cite{englund2012novelaproach}, the authors define a proximity-based kernel (PBK) that accounts for the number of branches between observations in each decision tree, defining the proximity between $i$ and $j$ as $p_{PBK}(i, j) = \frac{1}{T}\sum_{t = 1}^T\frac{1}{e^{w\cdot g_{ijt}}}$, where $T$ is the number of trees in the forest, $w$ is a user-defined parameter, and $g_{ijt}$ is the number of branches between observations $i$ and $j$ in tree $t$. $g$ is defined to be 0 if the observations reside in the same terminal node. The proximity quality was quantified using the classification accuracy when applied as a kernel in a support-vector machine. This definition showed some improvement over the original definition (Definition \ref{def:training_prox}) only when considering very small numbers of trees (5 or 10). It should be noted that this kernel is an attempt to improve random-forest-based similarity measures and not necessarily to explain the random forest. However, PBK is computationally expensive as all pair-wise branch distances must be computed within each tree.  This is not an issue for small numbers of trees, but typically the number of trees in a random forest is measured in hundreds (\texttt{randomForest}~\cite{randomForest2021} has a default of 500 trees). Additionally, this method adds a user-defined, tunable parameter which adds to its complexity.

The authors in \cite{davies2014randomforestkernel} describe an approach for computing random forest proximities in the context of a larger class of Random Partitioning Kernels. While most random forest proximities are determined primarily through associations within terminal nodes, this approach selects a random tree height and partitions the data based on this higher-level splitting. The authors do not compare  with other proximity definitions nor do they frame their work in the context of random forest proximities, but they do compare this random forest kernel to other typical kernels (linear, RBF, etc.) using a log-likelihood test. The random forest kernel outperformed the others in most cases and the authors visually demonstrated their kernel using 2D PCA plots. The code for this approach is not publicly available.

Cao et al. introduced two random forest dissimilarity measures that were used in the context of multi-view classification~\cite{Cao2020novelRFMV}. The first measure (denoted RFDisNC) weights the proximity values by the proportion of correctly-classified observations within each node, accounting for both in- and out-of-bag observations. The second (RFDisIH) is based on instance hardness. Euclidean distances between observations at each terminal node are calculated using only feature variables that were used as splitting variables leading to the terminal node to avoid the curse of dimensionality. These distances are then used as weights as a part of the dissimilarity measure.  Given this distance, they use $k$-Disagreeing Neighbors (DN) in the formulation of the dissimilarity measure:
\[
d_{t}\left(\mathbf{x}_{i}, \mathbf{x}_{j}\right)=\left\{\begin{array}{ll}
k D N\left(\mathbf{x}_{j}\right), & \text { if } v_i(t)=v_j(t) \\
1, & \text { otherwise }
\end{array}\right.,
\]
where
\[
k D N\left(\mathbf{x}_{i}\right)=\frac{\left|\{\mathbf{x}_{j}: \mathbf{x}_{j} \in k N N\left(\mathbf{x}_{i}\right), y_{j} \neq y_{i}\right\}|}{k},
\]
where $kNN(\mathbf{x}_i)$ is the set of the $k$-nearest neighbors of $\mathbf{x}_i$ in the training data.

The use of $k$-DN gives a notion of difficulty in classifying a particular observation. In the multiview problem, the dissimilarities from different views are averaged before classification. The authors showed that RFDisIH performed better overall on classification tasks compared with other multi-view methods. However, RFDisIH was not compared with other random forest proximities in their commonly-used applications (e.g., visualization or imputation). A similarity measure can be constructed from RFDisIH as RFProxIH $=1-$ RFDisIH. RFProxIH (adapted from RFDisIH) is also not specifically intended to explain the random forest's learning but uses the partitioning space to attempt to improve a forest-derived similarity measure.


While these alternative definitions have shown promise in their respective applications, their connection to the data geometry learned by the random forest is not clear. In contrast, we present a new definition of random forest proximities that exactly characterizes the random forest performance. In short, our new proximities provide a similarity measure that keeps the integrity of the forest.

\begin{definition}[Random Forest-Geometry- and Accuracy-Preserving Proximities (RF-GAP)]
\label{def:new_prox}
Let $B(t)$ be the multiset of (potentially repeated) indices of bootstrap (in-bag) observations. 
We define  $J_i(t)$ to be the set of in-bag observations which share the terminal node with observation $i$ in tree $t$, or $J_i(t) = B(t)\cap v_i(t)$, and $J_i(t) \subset M_i(t)$, which is the multiset including in-bag repetitions. Let $c_j (t)$ be the multiplicity of the index $j$ in the bootstrap sample. Then, for given observations $i$ and $j$, their proximity measure is defined as:
\[p_{GAP}(i, j)=\dfrac{1}{\left|S_{i}\right|} \sum_{t \in S_{i}}  \frac{c_j(t) \cdot I\left(j \in J_{i}(t)\right)}{\left|M_{i}(t)\right|}.\]
That is, the proximity between an observation $i$, and an observation $j$ is constructed using a weighted average of the counts of $j$ with the reciprocal of the number of in-bag observations in each shared terminal node serving as weights.

\end{definition}






This proposed definition is, in part, inspired by the work of Lin and Jeon~\cite{lin2006adaptiveNN}, who showed that random forests behave as nearest-neighbor predictors with an adaptive bandwidth. To see this, first consider that for kernel methods, such as the SVM, the prediction for the $i$th observation can be written as 
\[
\hat{y}_i(p)=\sum_{j=1}^n p(i,j)y_j,
\]
where $p(i,j)$ is some (potentially weighted) kernel function. A kernel function induces an inner product, and thus defines a similarity or proximity measure. Hence, Lin and Jeon proposed to use a proximity measure derived from the random forest for $p$ as their kernel. 

The three main proximity measures defined here ($p_{Or},$ $p_{OOB}$, and $p_{GAP}$) can be obtained by weighting the in- and out-of-bag observations differently in the prediction of $\hat{y}_i$. $p_{Or}$ equally weights in- and out-of-bag examples. Because the nodes in each tree of a random forest are typically grown to purity, the proximity measure $p_{Or}$ overemphasizes class separation compared to the actual random forest when using the proximity weighted prediction $\hat{y}_{p_{Or}}$. In this sense, this proximity-weighted prediction is a biased estimate of the random forest prediction function. 

On the other hand, $p_{OOB}$ doesn't use any of the in-bag observations, which are required to produce the random forest predictions. Thus the OOB proximities are also biased in the sense that the proximity-weighted prediction does not match the random forest prediction function. This is true even for large sample sizes (see Figures~\ref{fig:sample_size} and~\ref{fig:sample_errors}). 

In contrast, $p_{GAP}$ appropriately weights both in-bag and OOB samples such that the proximity-weighted prediction matches the random forest OOB predictions for all sample sizes. Thus using $p_{GAP}$ in the proximity-weighted prediction gives an unbiased estimate of the random forest prediction function.

OOB samples are commonly used to provide an unbiased estimate of the forest's generalization error rate~\cite{Breiman2001randomforests}. Thus, we use the forest's OOB predictions to assess the quality of our proximities. We show in Section~\ref{sec:proofs} that the random forest OOB prediction (and thus the generalization error rate) is reproducible as a weighted sum (for regression) or a weighted-majority vote (for classification) using the proximities in Definition~\ref{def:new_prox} as weights. Additionally, RF-GAP proximity-weighted predictions of test points match those of the random forest. Thus, this definition characterizes the random forest's predictions, keeping intact the learned data geometry. Subsequently, applications using this proximity definition will provide results that are truer to the random forest from which the proximities are derived. 

We note, however, that the RF-GAP proximities do not form a proper similarity measure as they are not symmetric. This is easily addressed in applications requiring symmetry by averaging $p_{GAP}(i, j)$ and $p_{GAP}(j, i)$. This was done in our experiments involving multi-dimensional scaling. In Figure~\ref{fig:symmetry}, we also demonstrate that the RF-GAP proximities become more symmetric as the number of trees increases.

\section{Random Forests as Proximity-Weighted Predictors}\label{sec:proofs}
Here we show that the random forest OOB and test predictions are reproducible using RF-GAP proximities as weights in a weighted sum (for regression) or a weighted-majority vote (for classification). We first show that for a given observation, the proximities are non-negative and sum to one. 

Note that for each observation, $i$, the RF-GAP definition assigns the value $p_{GAP}(i, i) = 0$. This is important when using the proximities as a predictor as an observation should not be used in the prediction of its own label. However, as a similarity measure, setting $p_{GAP}(i, i) = 0$ is not meaningful. Thus, for other proximity-based applications, we construct the main diagonal entries of the RF-GAP proximity matrix by passing through the random forest a duplicate, identical observation assigned out-of-bag status. 


\begin{proposition}\label{prop:weights}
Defining $p_{GAP}(i, i)= 0$, the RF-GAP proximities are non-negative and $\sum_{j=1}^N p_{GAP}(i, j)=1$.
\end{proposition}

\begin{IEEEproof}
It is clear from the definition that $p_{GAP}(i, j) \ge 0$ for all $i$, $j$.  The sum-to-one property falls directly from the definition:

\begin{align*}
    \sum_{j=1}^{N} p_{GAP}(i, j) &=\sum_{j=1}^{N} \frac{1}{\left|S_{i}\right|} \sum_{t \in S_{i}} \frac{c_j(t) \cdot I\left(j \in J_{i}(t)\right)}{\left|M_{i}(t)\right|} \\
    &=\frac{1}{\left|S_{i}\right|} \sum_{t \in S_{i}} \frac{1}{\left|M_{i}(t)\right|} \sum_{j=1}^{N} c_j(t) \cdot I{\left(j \in J_{i}(t)\right)} \\
    &=\frac{1}{\left|S_{i}\right|} \sum_{t \in S_{i}} \frac{1}{\left|M_{i}(t)\right|}\left|M_{i}(t)\right| \\
    &=\frac{1}{\left|S_{i}\right|} \sum_{t \in S_{i}} 1 \\
    &=1.
\end{align*}
\end{IEEEproof}

This proposition allows us to directly use RF-GAP as weights for classification or regression. In contrast, other proximities must be row-normalized to serve as weights. We show that the RF-GAP proximity-weighted prediction matches the random forest OOB prediction, giving the same OOB error rate in both the classification and regression settings. The OOB error rate is typically used to estimate the forest's generalization error and quantify its goodness of fit, this indicates that RF-GAP accurately represents the geometry learned by the random forest.

\begin{theorem}[Proximity-Weighted Regression]\label{thm:regression}
For a given training data set $\mathcal{S} = \left\{\left(\mathbf{x}_{1}, y_{1}\right) \ldots \left(\mathbf{x}_{N}, y_{N}\right) \right\}$, with $y_i \in \mathbb{R}$, the random forest OOB regression prediction can be reconstructed using the proximity-weighted sum with RF-GAP proximities as weights (Definition \ref{def:new_prox}). \end{theorem}

\begin{IEEEproof}
For a given tree, $t$, and $i\in O(t)$, the decision tree predictor of $y_i$ is the mean response of the in-bag observations in the appropriate terminal node.  That is, 

\[
    \hat{y}_{i}(t)=\frac{1}{\left|M_{i}(t)\right|} \sum_{j \in J_{i}(t)} c_j(t) \cdot y_{j} 
\]

The random forest prediction, $\hat{y}_i$, is the mean response over all trees for which $i$ is out of bag. That is,

\begin{align*}
    \hat{y}_{i} &=\frac{1}{\left|S_{i}\right|} \sum_{t \in S_{i}} \hat{y}_{i}(t) \\
    &=\frac{1}{\left|S_{i}\right|} \sum_{t \in S_{i}} \frac{1}{\left|M_{i}(t)\right|} \sum_{j \in J_{i}(t)} c_j(t) \cdot y_{j}.
\end{align*}

The proximity-weighted predictor, $\hat{y}_{i}^{p}$, is simply the weighted sum of responses, $\left\{y_j\right\}_{j \ne i}$. 

\begin{align*}
    \hat{y}_{i}^{p} &=\sum_{j=1}^{N} p_{GAP}(i, j) y_{j} \\
    &=\sum_{j=1}^{N}\left\{\frac{1}{\left|S_{i}\right|} \sum_{t \in S_{i}} \frac{c_j(t) \cdot I{\left(j \in J_{i}(t)\right)}}{\left|M_{i}(t)\right|}\right\} y_{j} \\
    &=\frac{1}{\left|S_{i}\right|} \sum_{t \in S_{i}} \frac{1}{\left|M_{i}(t)\right|} \sum_{j=1}^{N} c_j(t) \cdot I{\left(j \in J_{i}(t)\right)} y_{j} \\
    &=\frac{1}{\left|S_{i}\right|} \sum_{t \in S_{i}} \frac{1}{\left|M_{i}(t)\right|} \sum_{j \in J_{i}(t)} c_j(t) \cdot y_{j} \\
    &=\hat{y}_{i}.
\end{align*}
\end{IEEEproof}

\begin{theorem}[Proximity-Weighted Classification]\label{thm:classification}
For a given training data set $\mathcal{S}=\left\{\left(\mathbf{x}_{1}, y_{1}\right) \ldots \left(\mathbf{x}_{N}, y_{N}\right) \right\}$, with $y_i\in \{1, \cdots, K\}$ for all $i\in \{1,\cdots,N\}$, the random forest OOB classification prediction can be reconstructed by a weighted-majority vote using RF-GAP proximities (Definition \ref{def:new_prox}) as weights.
\end{theorem}

\begin{IEEEproof}
Given the training set $\left\{\left(\mathbf{x}_{1}, y_{1}\right) \ldots \left(\mathbf{x}_{N}, y_{N}\right) \right\}$, with $y_i\in \{1\ldots K\}$, for a given tree $t$ and observation $i\in O(t)$, the decision tree prediction of the label $y_i$ is determined by the majority vote among in-bag observations within the shared terminal node:

\[
\hat{y}_{i}(t)=\argmax _{l=1, \ldots, K} \sum_{j \in J_{i}(t)} c_j(t) \cdot I{\left(y_{j}=l\right)}.
\]

Thus, the random forest classification prediction, $\hat{y}_i$, is the most popular predicted class across all $t\in S_i$:

{\footnotesize
\begin{align*}
    \hat{y}_{i} &=\argmax _{k=1, \ldots, K} \sum_{t \in S_{i}} I{\left(\hat{y}_{i}(t)=k\right)} \\
    &=\argmax _{k=1, \ldots, K} \sum_{t \in S_{i}} I\left(\left\{\argmax _{l=1, \ldots, K} \sum_{j \in J_{i}(t)} c_j(t) \cdot I{\left(y_{j}=l\right)}\right\}=k\right)
\end{align*}
}%

We show equivalence with the proximity-weighted predictor. The proximity-weighted predictor predicts the class with the largest proximity-weighted vote:

{\footnotesize
\begin{align*}
    \hat{y}_{i}^{p} &=\argmax _{k=1, \ldots, K} \sum_{j=1}^{N} p_{GAP}(i, j) I{\left(y_{j}=k\right)} \\
    &=\argmax _{k=1, \ldots, K}\left\{\sum_{j=1}^{N}\left(\frac{1}{\left|S_{i}\right|} \sum_{t \in S_{i}} \frac{c_j(t) \cdot I{\left(j \in J_{i}(t)\right)}}{\left|M_{i}(t)\right|}\right) I{\left(y_{j}=k\right)}\right\} \\
    &=\argmax _{k=1, \ldots, K}\left\{\frac{1}{\left|S_{i}\right|} \sum_{t \in S_{i}} \frac{1}{\left|M_{i}(t)\right|}\left(\sum_{j=1}^{N} c_j(t) \cdot I{\left(j \in J_{i}(t), y_{j}=k\right)}\right)\right\} \\
    &=\argmax _{k=1, \ldots, K}\left\{\frac{1}{\left|S_{i}\right|} \sum_{t \in S_{i}} \frac{1}{\left|M_{i}(t)\right|} \sum_{j \in J_{i}(t)} c_j(t) \cdot I{\left(y_{j}=k\right)}\right\} \\
    &=\argmax _{k=1, \ldots, K}\left\{\sum_{t \in S_{i}} \frac{1}{\left|M_{i}(t)\right|} \sum_{j \in J_{i}(t)} c_j(t) \cdot I{\left(y_{j}=k\right)}\right\}.
\end{align*}
}%

The last line holds as $|S_i|$ does not depend on $k$ and just rescales the summations. As classification trees in a random forest are grown until terminal (leaf) nodes are pure, all in-bag observations belong to the same class.  Denote the common class for any observation $j \in M_i(t)$ as $y_{i, t}$. Then the single tree predictor is given by 

\begin{align*}
    \hat{y}_{i}(t) &=\argmax _{l=1, \ldots, K} \sum_{j \in J_{i}(t)} c_j(t) \cdot I{\left(y_{j}=l\right)} \\
    &=y_{i, t}.
\end{align*}

and the random forest predictor is 

\[
\hat{y}_{i}=\argmax _{k=1, \ldots, K} \sum_{t \in S_{i}} I{\left(y_{i, t}=k\right)}.
\]

The proximity-weighted predictor is thus

{\footnotesize
\begin{align*}
    \hat{y}_{i}^{p} &=\argmax _{k=1, \ldots, K}\left\{\sum_{t \in S_{i}} \frac{1}{\left|M_{i}(t)\right|} \sum_{j \in J_{i}(t)} c_j(t) \cdot I{\left(y_{j}=k\right)}\right\} \\
    &=\argmax _{k=1, \ldots, K}\left\{\sum_{t \in S_{i}} \frac{1}{\left|M_{i}(t)\right|} \sum_{j \in J_{i}(t)} c_j(t) \cdot I{\left(y_{i, t}=k\right)}\right\} \\
    &=\argmax _{k=1, \ldots, K}\left\{\sum_{t \in S_{i}} \frac{1}{\left|M_{i}(t)\right|} \left|M_{i}(t)\right| I{\left(y_{i, t}=k\right)}\right\} \\
    &=\argmax _{k=1, \ldots, K} \sum_{t \in S_{i}} I{\left(y_{i, t}=k\right)} \\
    &=\hat{y}_{i}.
\end{align*}
}%

\end{IEEEproof}

\begin{figure}[htb!]
    \centering
    \includegraphics[width = .5\textwidth]{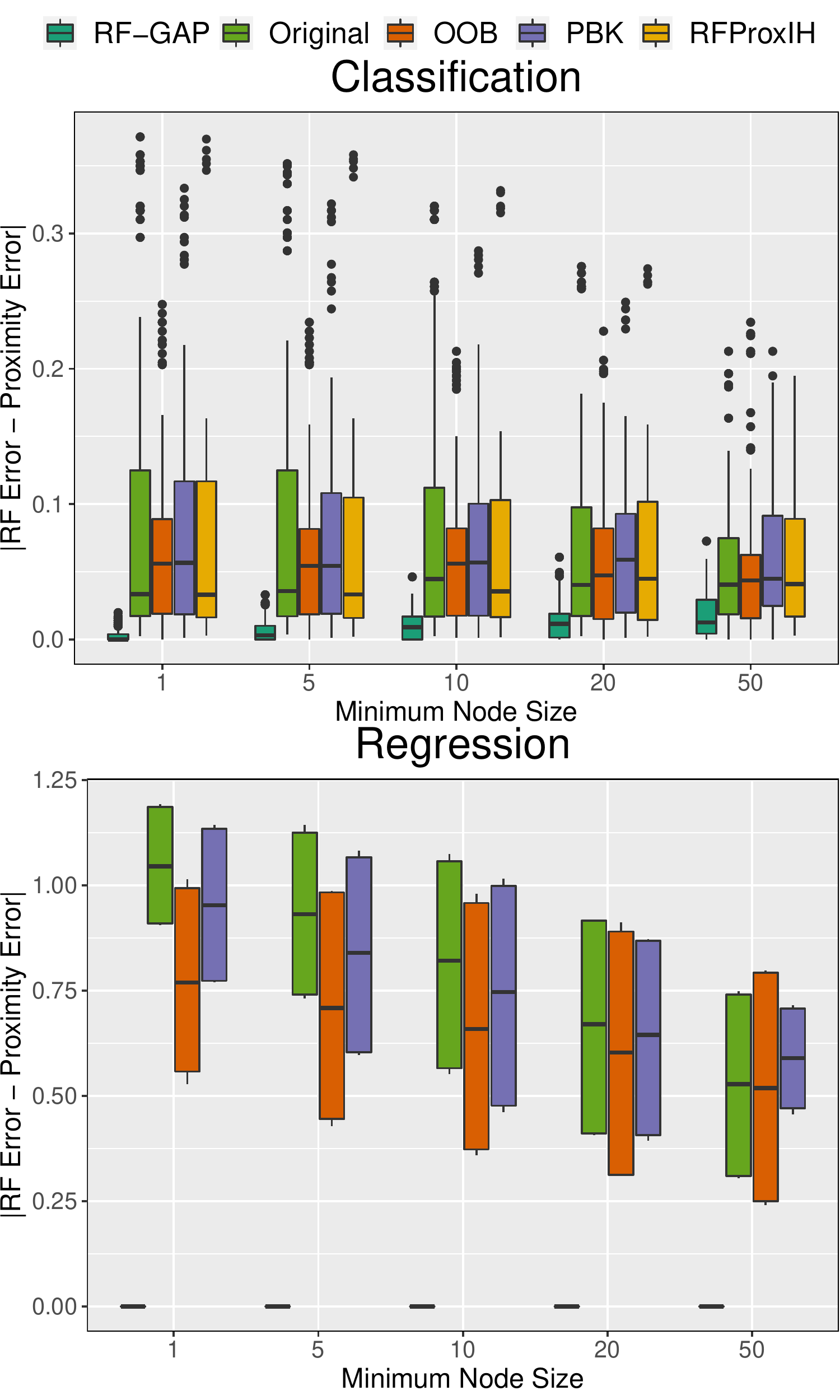}
    \caption{Random forest predictions are compared here with the proximity-weighted predictions for both regression and classification problems. For each of the datasets listed in supplemental Table~\ref{tab:datasets}, the differences were calculated across five random seeds. For classification, the RF-GAP prediction variations tend to increase with node size, although they are still more similar to the RF predictions when compared with the other proximities. For regression, RF-GAP predictions perfectly match the RF predictions regardless of node size. Note that RFProxIH is incompatible with regression tasks.}
    \label{fig:min_node_size}
\end{figure}

We note that the proof for the proximity-weighted classification problem requires that the trees in the forest are grown until the terminal nodes are pure. We show empirically, however, that RF-GAP proximity predictions more closely match the random forest's predictions than the other proximities even when trees are pruned.  See Figure~\ref{fig:min_node_size} for a comparison. Here we compared proximity-weighted predictions using minimum node sizes of 1, 5, 10, 20, and 50. For regression problems, RF-GAP proximity predictions match those of the random forest perfectly, while this isn't always the case for classification. The difference increases with node size as ties become more likely. We also note that the random forest and proximity-weighted predictions produce higher error rates with the increase in minimum node size (Figure~\ref{fig:node_size_error}).

\section{Experimental Validation of Proximity-Weighed Prediction}\label{sec:results}


To demonstrate that the RF-GAP proximities preserve the random-forest learned data geometry, we empirically validate Theorems~\ref{thm:regression} and \ref{thm:classification} from Section~\ref{sec:proofs}, demonstrating that the random forest predictions are preserved in the proximity construction. We also compare to the proximity-weighted predictor using the original random forest proximity definition (Definition~\ref{def:training_prox}), the OOB adaptation (Definition~\ref{def:oob_prox}), PBK~\cite{englund2012novelaproach}, and RFProxIH~\cite{Cao2020novelRFMV}.

We compared proximity-prediction results on 24 datasets as described in Table~\ref{tab:datasets}. Each dataset was randomly partitioned into training (70\%) and test (30\%) sets. For each dataset, the same trained random forest was used to produce all compared proximities. Tables~\ref{tab:test_differences} and \ref{tab:training_differences} show the differences between the random forest predictions and the proximity-weighted predictions for test and training sets, respectively. The proximity-weighted error rates using RF-GAP proximities nearly exactly match the random forest OOB error rates; discrepancies are due to random tie-breaking. In contrast, the original definition, PBK, and RFProxIH typically produce much lower training error rates, suggesting overfitting. The OOB proximities (Definition~\ref{def:oob_prox}) produce training error rates that are sometimes lower and sometimes higher than the random forest OOB error rate. 

\begin{figure}[t!]
    \centering
    \includegraphics[width = .45\textwidth]{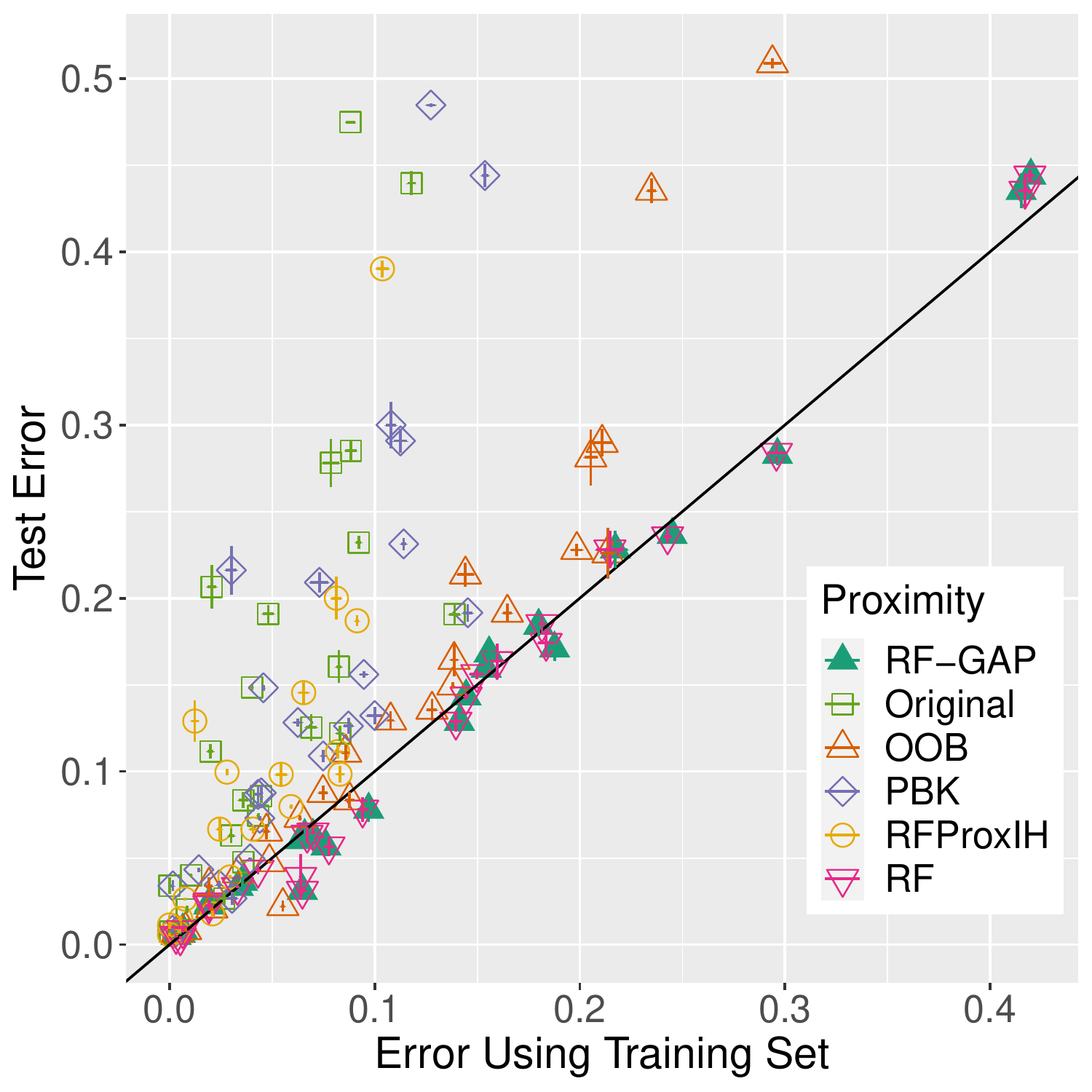}
    \caption{Training vs. test error of the proximity-weighted predictions 24 datasets described in the supplemental Table~\ref{tab:datasets}. Errors were averaged across five runs. We see that the original proximities, PBK, and RFProxIH, tend to overfit the training data, as demonstrated by points above the line $y = x$. The random forest errors and RF-GAP nearly perfectly align in most cases and are each well-described by the line.}
    \label{fig:train_test_scatter}
\end{figure}

\begin{figure}[htb!]
    \centering
    \includegraphics[width = .5\textwidth]{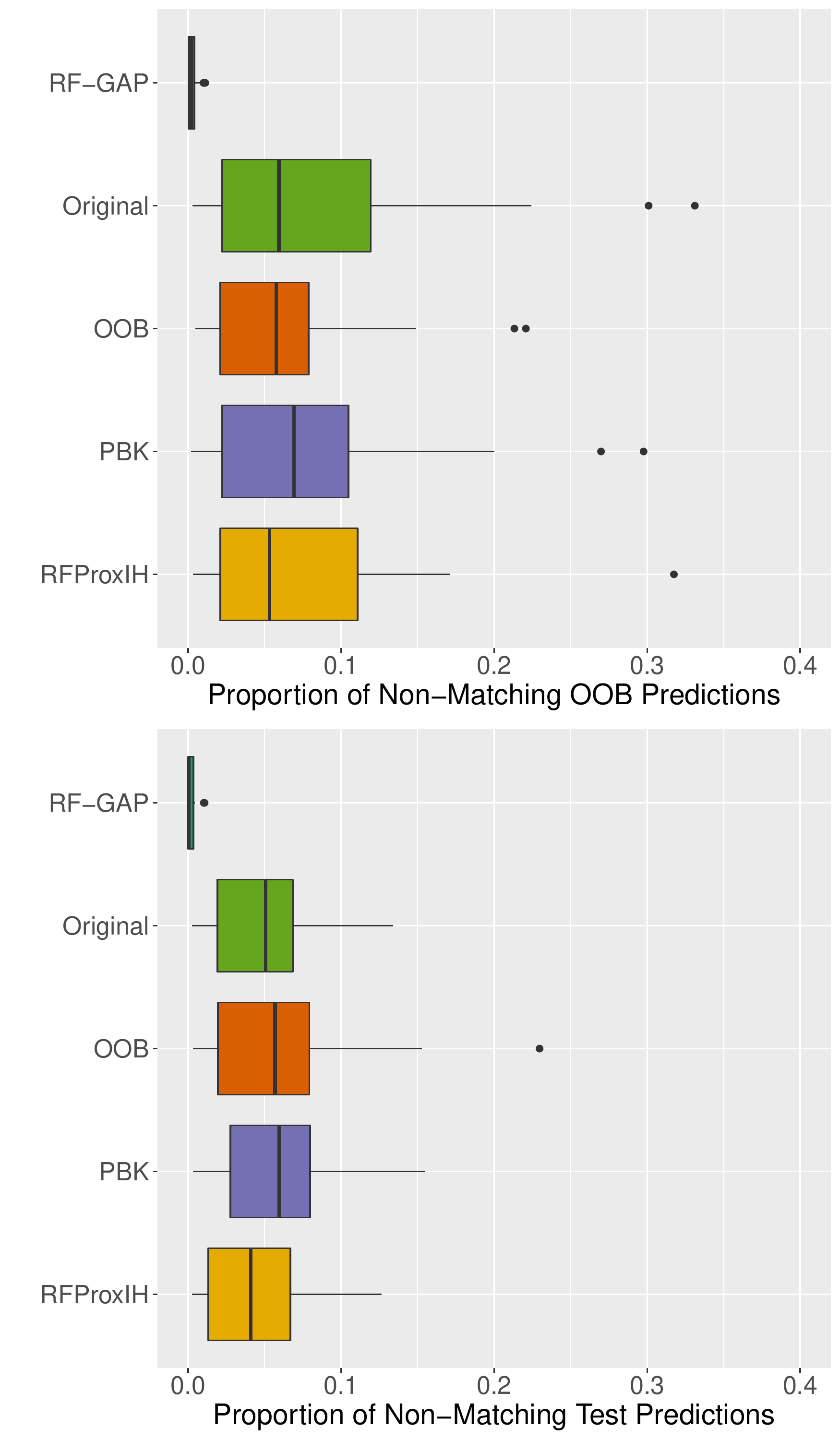}
    \caption{These boxplots show the proportion of proximity-weighted OOB and test predictions which do not match the respective random forest predictions. Proximity-based training predictions were compared with the RF out-of-bag errors. RF-GAP proximity predictions most nearly match the random forest predictions for both the training (top) and test (bottom) data, thus, best preserving the geometry learned by the random forest. Datasets were split into 70\% training and 30\% test data. Errors were averaged across 5 runs. The dataset details can be found in Table~\ref{tab:datasets}.}
    \label{fig:train_test_boxplots}
\end{figure}

The RF-GAP proximities also generally produce the lowest test errors. This can be seen in Figure~\ref{fig:train_test_scatter}, which plots the training versus test error rates using the different proximity measures.  Table~\ref{tab:slopes} gives the regression slope for each proximity definition. From here it is clear that the original proximities, PBK, and RFProxIH overfit the training data. This is corroborated in Figure~\ref{fig:train_test_boxplots}, which plots the difference between the random forest out-of-bag error rates and the proximity-weighted errors across the same datasets, demonstrating that the RF-GAP predictions nearly perfectly match those of the random forest for both training and tests sets. It is clear that the original definition, PBK, and RFProxIH generally overfit the training data in contrast.   We re-emphasize that PBK and RFProxIH were not intended to demonstrate the forest's learning but to use the forest as a means to construct a similarity measure.

It is well-established that increasing the number of trees increases the random forest generalization error rate for standard measures of error~\cite{Breiman2001randomforests, probst2017tuneornot}, while the typical random forest default of 500 trees is generally sufficient~\cite{wright2017ranger, randomForest2021}. We compared proximity-based predictions for smaller numbers of trees (5, 10, 50, 100, 250) and demonstrated that RF-GAP predictions still most accurately match those of the random forest regardless of forest size. However, for classification problems, tie breaks are more often required with a small number of trees. Results can be found in Figure~\ref{fig:num_trees}.

In addition to comparing proximity-weighted predictions to the random forest predictions, we compared predictions across the proximity definitions. Generally, RF-GAP predictions differed the most from the other proximity-weighted predictions, especially when using the training set. This is not surprising as the RF-GAP proximities are defined to match the random forest OOB predictions. See more details in Figure~\ref{fig:test_agreement}.

\begin{table}[ht!]
    \centering
    \caption{The regression slopes of each proximity type corresponding to the points in Figure~\ref{fig:train_test_scatter}. RF-GAP does not exhibit bias towards the training data as they have a slope close to one, while the larger slope of the other proximity definitions indicates they are overfitting the training data.}
    \begin{tabular}{|l|c|}
    \hline
       Type  &  Slope\\
       \hline
        RF-GAP   & 1.040\\
        Original & 2.434\\
        OOB      & 1.543\\
        PBK      & 2.287\\
        RFProxIH & 2.147\\
        \hline
    \end{tabular}
    \label{tab:slopes}
\end{table}

\section{Comparison of Common Proximity-based Applications}

In this section, we demonstrate that using RF-GAP in multiple applications leads to improved performance relative to the other random forest proximity measures. We compare scatterplots generated by applying MDS to the proximities for visualization in Section~\ref{subsec:MDS}, data imputation in Section~\ref{subsec:imputation}, and outlier detection in Section~\ref{subsec:outliers}. Random forest proximities have already been established as successful in these applications (see Section~\ref{sec:applications}). Thus, we focus our comparisons in these applications on existing random forest proximity definitions to show that the improved representation of the random forest geometry in RF-GAP leads to improved performance. In each comparative experiment, we train the random forests using default hyperparameters. Each set of proximities is constructed using the same forest to ensure differences are attributed entirely to the proximity definition and not to the inherent randomness of the forest. Additional results beyond those presented here are given in Appendix~\ref{sec:appendix}.

\subsection{Visualization using Multi-Dimensional Scaling}\label{subsec:MDS}

 Leo Breiman~\cite{Breiman2001randomforests, BreimanRandomForest:Online} first introduced the use of multi-dimensional scaling on random forest proximities to visualize the forest's learning. Since then, MDS with random forest proximities has been used in many applications including tumor profiling~\cite{Shi2005tumor}, visualizing biomarkers~\cite{finehout2007cerebrospinal}, pathway analysis~\cite{pang2006pathways}, multi-view learning~\cite{gray2011dementia, cao2019multi-viewradiomics}, and unsupervised learning~\cite{horvath2006unsupervised}. Here, we compare visualizations using the five different proximities. In each case, metric MDS was applied to $\sqrt{1 - \text{prox}}$ to produce two-dimensional visualizations. 
 
 \begin{figure*}[htb!]
    \centering
    \includegraphics[width = \textwidth]{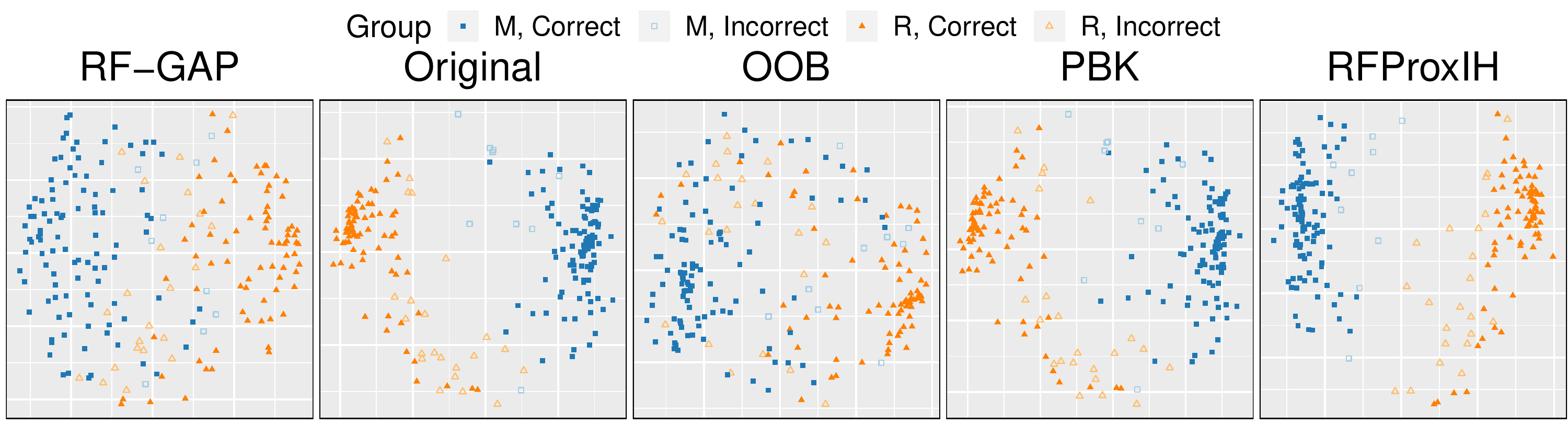}
    \caption{Here is a comparison of MDS embeddings using different RF proximity definitions. Proximities were constructed from a random forest trained on the two-class Sonar dataset (208 observations of 60 variables) from the UCI repository which gave an OOB error rate of 14.9\%. Multi-dimensional scaling (MDS) was applied to $\sqrt{1 - \text{prox}}$ using RF-GAP proximities, the original proximities, OOB proximities (Definition \ref{def:oob_prox}), PBK~\cite{englund2012novelaproach}, and RFProxIH~\cite{Cao2020novelRFMV}. Using RF-GAP proximities, the visualization depicts a good representation of the forest's classification problem. For correctly-classified points (dots), there are two clear groupings, while misclassified points (squares) are generally located between the groupings or found within the opposite class cluster, albeit closer to the decision boundary than not. The original definition, PBK, and RFProxIH  over-exaggerate the separation between classes. This is apparent in examples in the figure as the two classes appear nearly linearly separable which does not accurately depict the random forest's performance on the dataset. Using only OOB samples to generate the proximities improves upon those three but seems to add some noise to the visualization. There are still two major class clusters, but some correctly classified points are found farther inside the opposite class' cluster compared to the RF-GAP visualization. More examples can be found in Appendix~\ref{subsec:mds_outlier}.}
    \label{fig:sonar}
\end{figure*}
 
 Figure~\ref{fig:sonar} gives an example of MDS applied to the classic Sonar dataset from UCI~\cite{UCI2019} with an OOB error rate was 14.9\%. RF-GAP proximities (Figure~\ref{fig:sonar}) show two class groupings with misclassified observations between the groups or within the opposing class. The original proximities, PBK, and RFProxIH show a fairly clear separation between the two classes. For these proximities, they appear nearly linearly separable which does not accurately reflect the data nor the geometry learned by the random forest given an error rate of nearly 15\%. Definition \ref{def:oob_prox} has a similar effect as the RF-GAP definition, but with a less clear boundary and seemingly misplaced observations that are deep within the wrong class. These results suggest that RF-GAP can lead to improved supervised visualization and dimensionality reduction techniques. See Appendix~\ref{subsec:mds_outlier} for similar experiments.

\subsection{Imputation}\label{subsec:imputation}

In~\cite{Kokla2019metabolomicsimputation}, experimental results showed that, out of nine compared imputation methods across seven different missing mechanisms (missing at random, missing completely at random, etc.), random forest-based imputation was generally the most accurate. This has been corroborated by~\cite{pantanowitz2009missingdata, Shah2014caliber, Ramosaj2019predictmiss}. Here we describe the algorithm for random forest imputation and compare results using various proximity definitions. 

To impute missing values of variable $j$: \begin{enumerate}[label = (\arabic*)]
    \item If variable $j$ is continuous, initialize the imputation with the median of the in-class values of variable $j$; otherwise, initialize it with the most frequent in-class value. In the regression context, the median or most frequent values are computed across all observations without missing values in variable $j$.
    \item Train a random forest using the imputed dataset and
    \item construct the proximity matrix from the forest.
    \item If variable $j$ is continuous, replace the missing values with the proximity-weighted sum of the non-missing values. If categorical, replace the missing values with the proximity-weighed majority vote.
    \item Repeat steps 2 - 4 as required.  In many cases, a single iteration is sufficient.
\end{enumerate}

\begin{figure*}
    \centering
    \includegraphics[width = \textwidth]{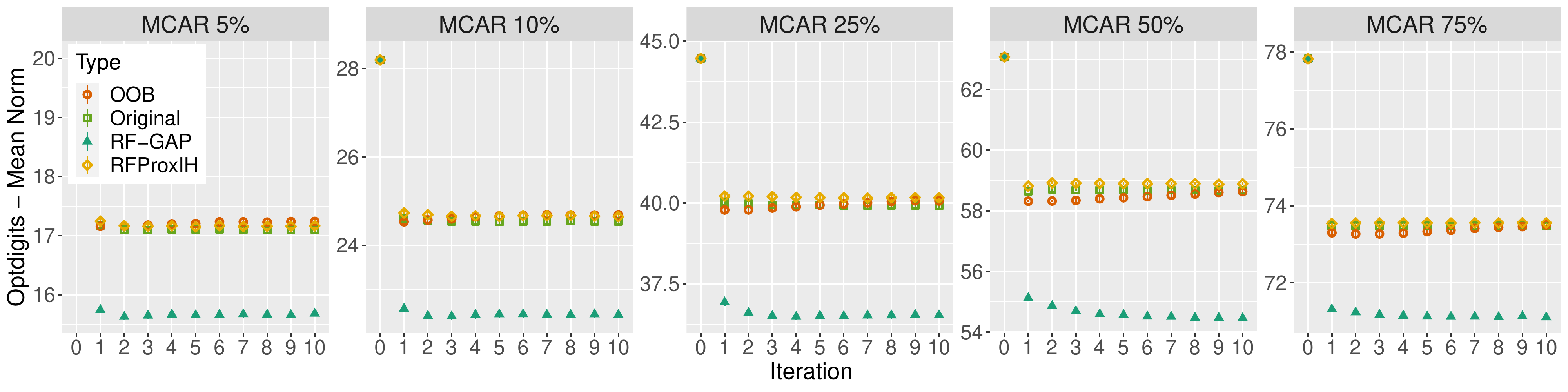}
    \caption{Imputation of values MCAR of the optical digits dataset~\cite{UCI2019}. The experiment was repeated over 5 runs and across 10 iterations. RF-GAP proximities produced the most accurate imputations.}
    \label{fig:optdigits_imp}
\end{figure*}

We show empirically that random forest imputation is generally improved using the RF-GAP proximity definition. For our experiment, we selected datasets that are described in Table~\ref{tab:datasets} and for each dataset, we removed 5\%, 10\%, 25\%, 50\%, and 75\% of values at random (that is, the values are missing completely at random, or MCAR), using the \texttt{missMethods}~\cite{rockel2022missMethods} \texttt{R} package. Two comparisons were made: 1) we computed the mean MSE across 100 repetitions using a single iteration, and 2) we computed the mean MSE (across 5 repetitions) at each of 10 iterations. An example using 10 iterations is given in Figure~\ref{fig:optdigits_imp}.

A summary of performance rankings is given in Table~\ref{tab:imp_ranks}. Across all compared proximity definitions (RF-GAP, OOB, original, and RFProxIH), RF-GAP achieved the best imputation rankings across all percentages of missing values.  For full results, see Table~\ref{tab:imp} which gives the average MSE across 100 repetitions using a single iteration of the above-described algorithm. The number of observations and variables for each dataset are also provided in the table. 

\begin{table}[htb!]
\caption{The average ranks of the imputation scores across the various UCI datasets and various percentages of missing values. Each imputation experiment was repeated 100 times with different random initialization. RF-GAP generally produces the best imputations across all percentages of missing values. See Table~\ref{tab:imp} for the full imputation results.}

\label{tab:imp_ranks}
\centering
\begin{tabular}{|c|c|c|c|c|c|}
  \hline
& 5\% & 10\% & 25\% & 50\% & 75\% \\ 
  \hline
RF-GAP	&	1.10	&	1.20	&	1.15	&	1.35	&	1.65	\\
Original	&	2.55	&	2.70	&	2.65	&	2.60	&	2.35	\\
OOB	&	2.50	&	2.55	&	2.45	&	2.40	&	2.30	\\
RFProxIH	&	3.20	&	3.33	&	3.33	&	3.40	&	3.20	\\

   \hline
\end{tabular}
\end{table}

Figures~\ref{fig:imputation_first}, \ref{fig:imputation_second}, and \ref{fig:imputation_third} compare imputation results across 17 datasets~\footnote{Large datasets were not used as RFProxIH is computationally inefficient. Also, datasets already missing values were not used.}, using 10 iterations in each experiment with the mean MSE and standard errors recorded for each of the repetitions. The value recorded at iteration 0 is the MSE given the median- or majority-imputed datasets. In many cases, the imputation appears to converge quickly with relatively few iterations. Generally, the RF-GAP proximities outperformed the other definitions at each number of iterations. These results suggest that RF-GAP can be used to improve random forest imputation. For high percentages of missing values (at least 75\%), or for small datasets, the random forest imputation does not always converge and performance may actually decrease as the number of iterations increases.


\subsection{Outlier Detection}\label{subsec:outliers}

\begin{figure*}[htb!]
    \centering
    \includegraphics[width = \textwidth]{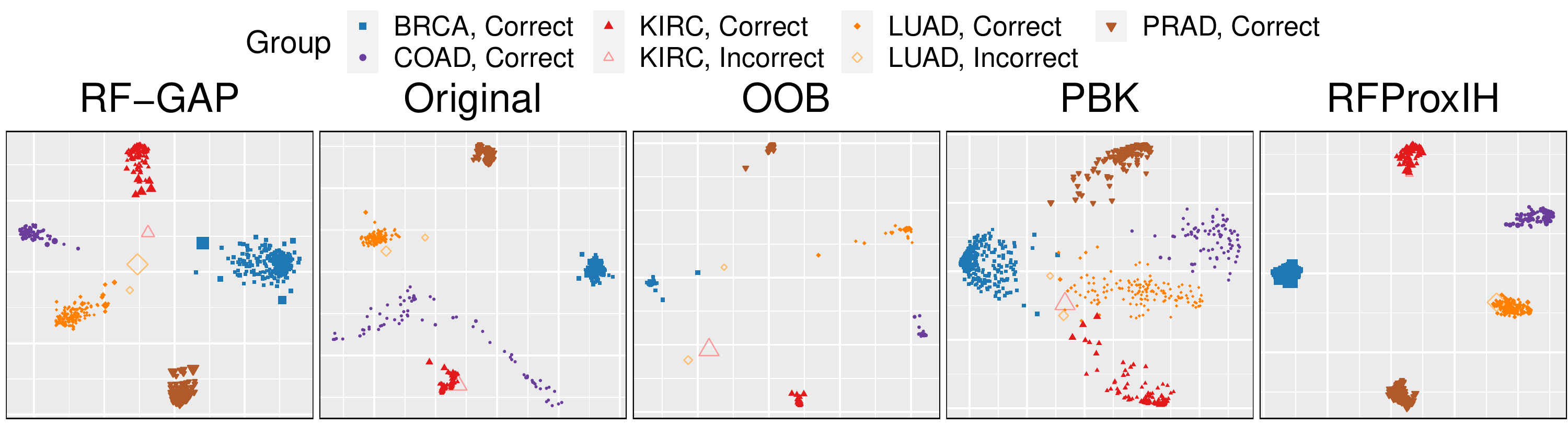}
    \caption{MDS applied to the random forest proximities computed from the Gene Expression Cancer dataset from UCI~\cite{UCI2019}. The point sizes are inversely proportional to the average proximity of a given observation to all other within-class observations. The misclassifications in RF-GAP, and OOB, are understandable based on the distance from their respective clusters. The original proximities and RFProxIH do not clearly account for the misclassified points. The outlier measure scaling in RF-GAP gives a clear reflection of the distance of points to their respective clusters.}
    \label{fig:outliers}
\end{figure*}

\begin{figure}[!ht]
    \centering
    \includegraphics[width = .48\textwidth]{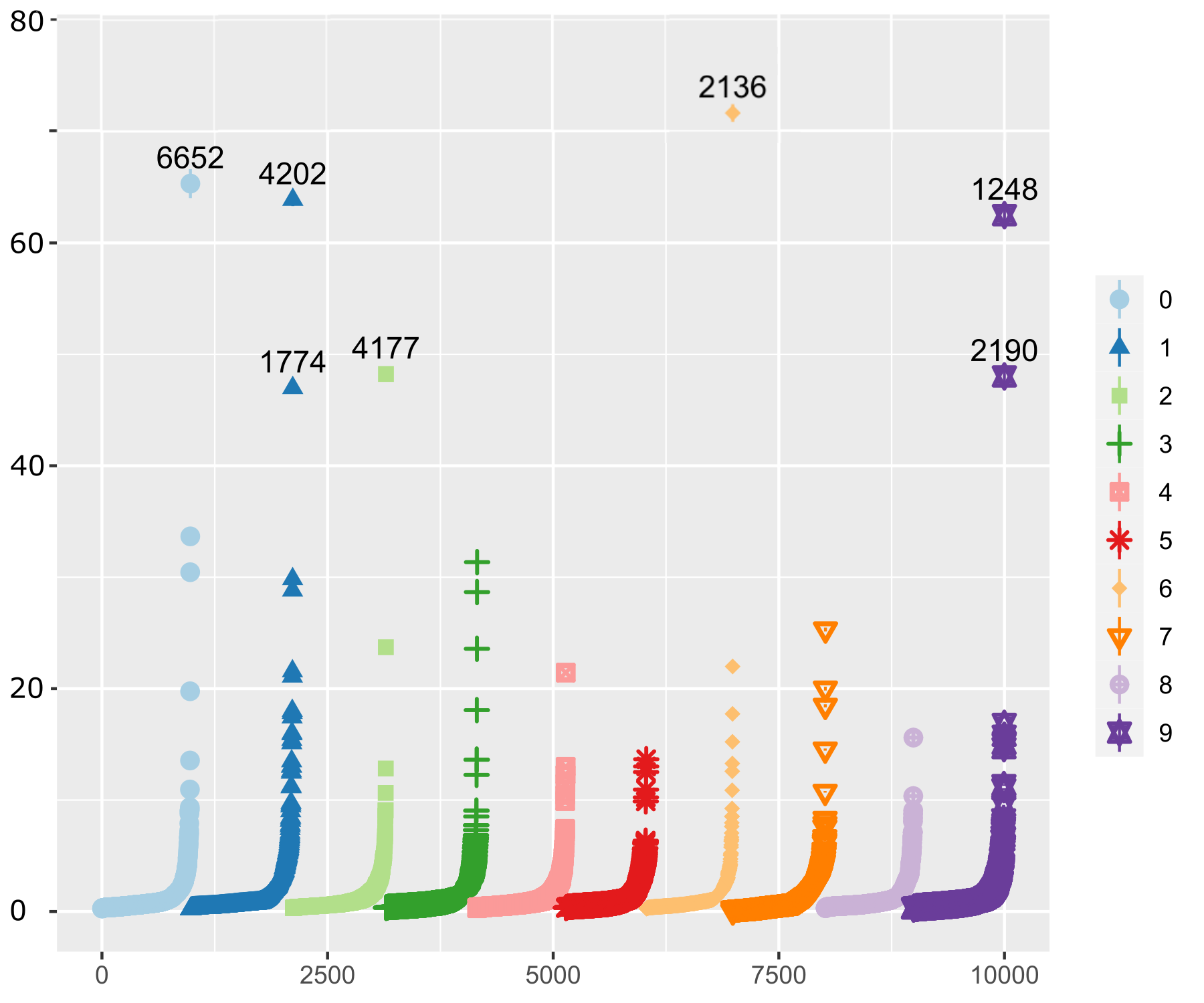}
    \caption{A sorted plot of the outlier measures for the MNIST dataset as provided by RF-GAP. The vertical axis is the outlier measure as described in Section~\ref{subsec:outliers}. The top seven outlying images are labeled with an index and shown in Figure~\ref{fig:mnist_examples}.}
    \label{fig:mnist_outliers}
\end{figure}

\begin{figure}[!ht]
    \centering
    \includegraphics[width = .48\textwidth]{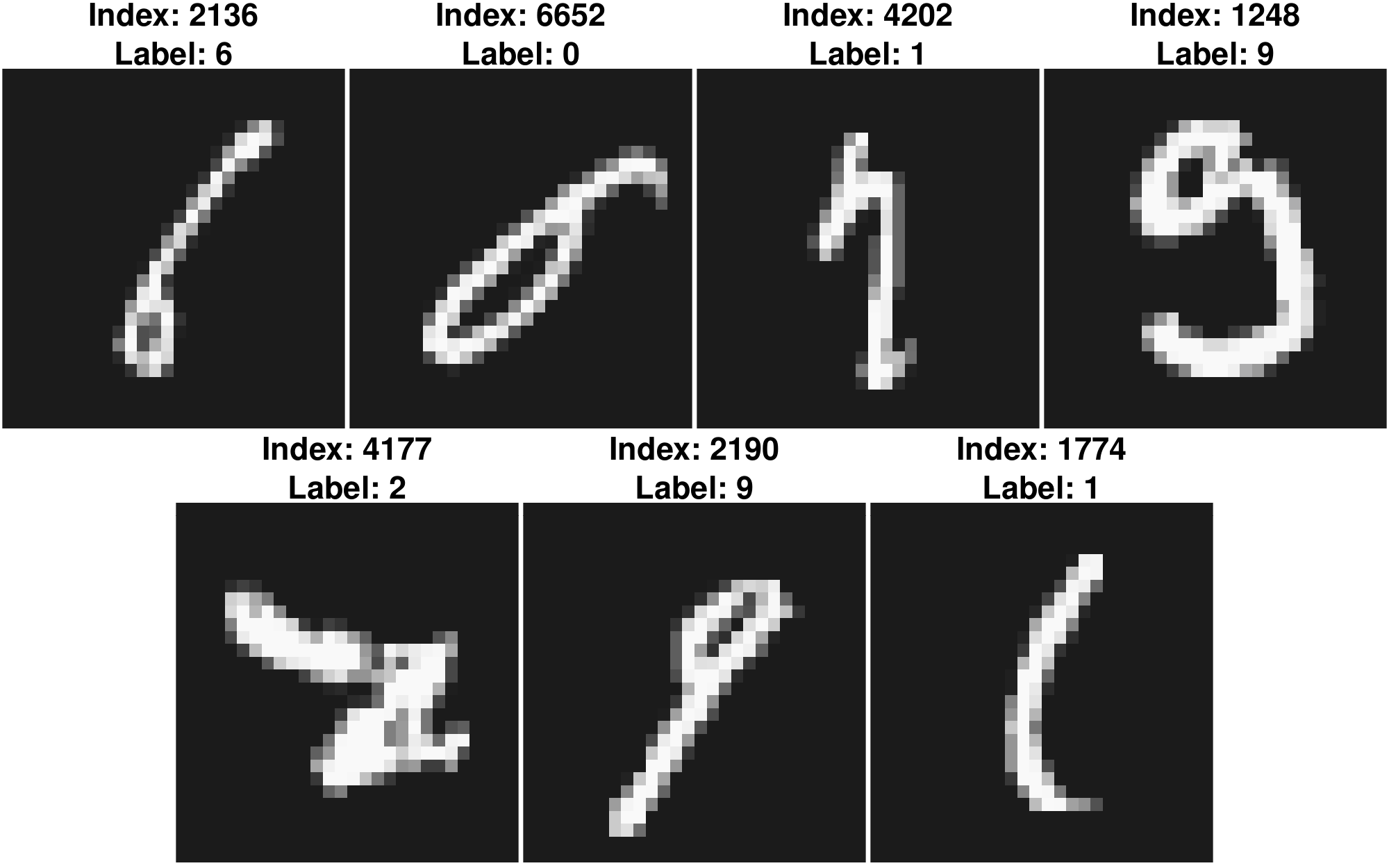}
    \caption{The top seven outlying digits per RF-GAP (see Figure~\ref{fig:mnist_outliers}). Some of these digits may even be difficult for a human to classify, corroborating the RF-GAP outlier score.}
    \label{fig:mnist_examples}
\end{figure}

Random forests can be used to detect outliers in the data. In the classification setting, outliers may be described as observations with measurements that significantly differ from those of other observations within the same class. In some cases, these outliers may be similar to observations in a different class, or perhaps they may distinguish themselves from observations in all known classes. In either case, outlying observations may negatively impact the training of many classification and regression algorithms, although random forests themselves are rather robust to outliers in the feature variables~\cite{Cutler2012randomforests}. 

Random forest proximity measures can be used to detect within-class outliers as outliers are likely to have small proximity measures with other observations within the same class.  Thus, small average within-class proximity values may be used as an outlier measure. We describe the algorithm as follows: \begin{enumerate}[label = (\arabic*)]
    \item For each observation, $i$, compute the raw outlier measure score as $\sum_{j\in \text{class}(i)}{\frac{n}{prox^2(i, j)}}$.
    
    \item Within each class, determine the median and mean absolute deviation of the raw scores.  
    
    \item Subtract the median from the raw score and divide the result by the mean absolute deviation.
\end{enumerate}

The outlier detection measure may be used in conjunction with MDS for visualization. See Figure~\ref{fig:outliers} for an example using the Gene Expression Cancer dataset from UCI~\cite{UCI2019} which has 5 classes across 801 observations and 20531 variables. Here, the point sizes of the scatter plot are proportionally scaled to the outlier measure. From the figure, it is clear that points outside of their respective class clusters have higher outlier measures. That the outlier measure is inversely proportional to the average proximity to within-class observations is clear in the case of RF-GAP proximities and is not very clear in the cases of the original definition and RFProxIH. This suggests that RF-GAP can be used to improve random forest outlier detection. 

Figures~\ref{fig:mnist_outliers} shows the RF-GAP outlier scores for the MNIST dataset. From this plot, outliers may be visually identified as the datapoints with substantially higher outlier scores compared to their class. Figure~\ref{fig:mnist_examples} shows some of the outlying images, some of which may be difficult for humans to distinguish.  Appendix~\ref{subsec:mds_outlier} contains further experiments.

\section{Conclusion}

In this paper, we presented a new definition of random forest proximities called RF-GAP that  characterizes the random forest out-of-bag prediction results using a weighted nearest neighbor predictor. We proved that the performance of the proximity-weighted predictor exactly matches the out-of-bag prediction results of the trained random forest and also demonstrated this relationship empirically. Thus, RF-GAP proximities capture the random forest-learned data geometry which provides empirical improvements over other existing definitions in applications such as proximity-weighted prediction, missing data imputation, outlier detection, and visualization. 

Additional random forest proximity applications can be explored in future works, including quantifying outlier detection performance, comparing with non-tree-based methods, assessing variable importance, and applying RF-GAP to multi-view learning. This latter application shows promise as classification accuracy was greatly increased after combining proximities in~\cite{gray2011dementia, cao2019multi-viewradiomics} using other definitions. An interesting visualization application was introduced in~\cite{rhodes2021rfphate} which may see improvements using geometry-preserving proximities. Multi-view learning may also be paired with this approach in some domains to visualize and assess contributions from the various modes or to perform manifold alignment. 

An additional area of improvement is scalability. Our approach is useful for datasets with a few thousand observations, but we can expand its capabilities by implementing a sparse version of RF-GAP. Further adaptations may make random forest proximity applications accessible for even larger datasets.


%



\ifCLASSOPTIONcaptionsoff
  \newpage
\fi

\bibliographystyle{IEEEtran}
\bibliography{references.bib}

\begin{IEEEbiography}[{\includegraphics[width=1in,height=1.25in,clip,keepaspectratio]{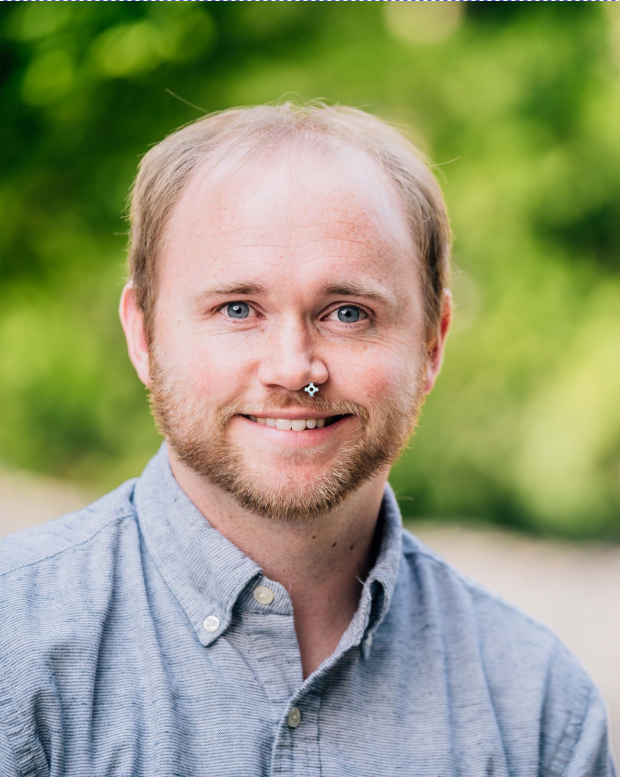}}]{Jake S. Rhodes} is an assistant professor at Idaho State University. He earned his Ph.D. at Utah State University (USU). He has a B.S. Degree in Applied Mathematics from Weber State University (2017) and a B.A. Degree in Finance and Economics from Utah State University (2015). During the last year of his doctoral program, he participated in the National Security Internship Program (NSIP) at Pacific Northwest National Laboratory, where he worked in areas of vision and robust deep-learning models. His research interests include machine learning (supervised and unsupervised), deep learning, and dimensionality reduction.

\end{IEEEbiography}

\begin{IEEEbiography}[{\includegraphics[width=1in,height=1.25in,clip,keepaspectratio]{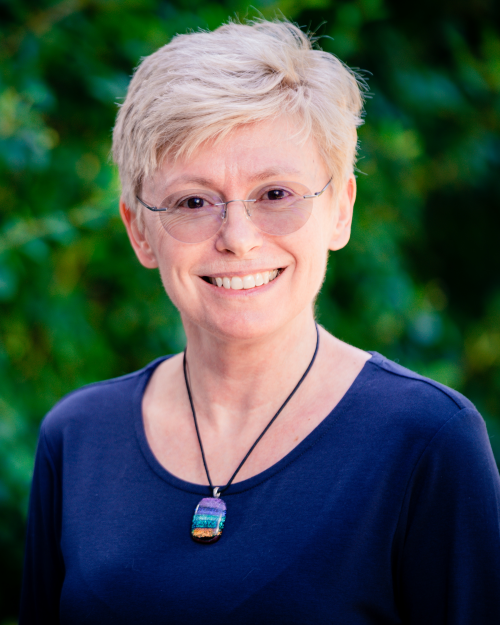}}]{Adele Cutler} is a professor emerita in the Department of Mathematics and Statistics at Utah State University (USU) and the Interim Associate Dean of Faculty and Research in the College of Science at USU. She holds a B.Sc. (Hons) degree in Mathematics from the University of Auckland and an M.S. and Ph.D. in Statistics from the University of California, Berkeley. She is one of the pioneering researchers in the development of random forests and archetypal analysis. Her research interests are in statistical learning, statistical computing, random forests, and archetypal analysis.

\end{IEEEbiography}

\begin{IEEEbiography}[{\includegraphics[width=1in,height=1.25in,clip,keepaspectratio]{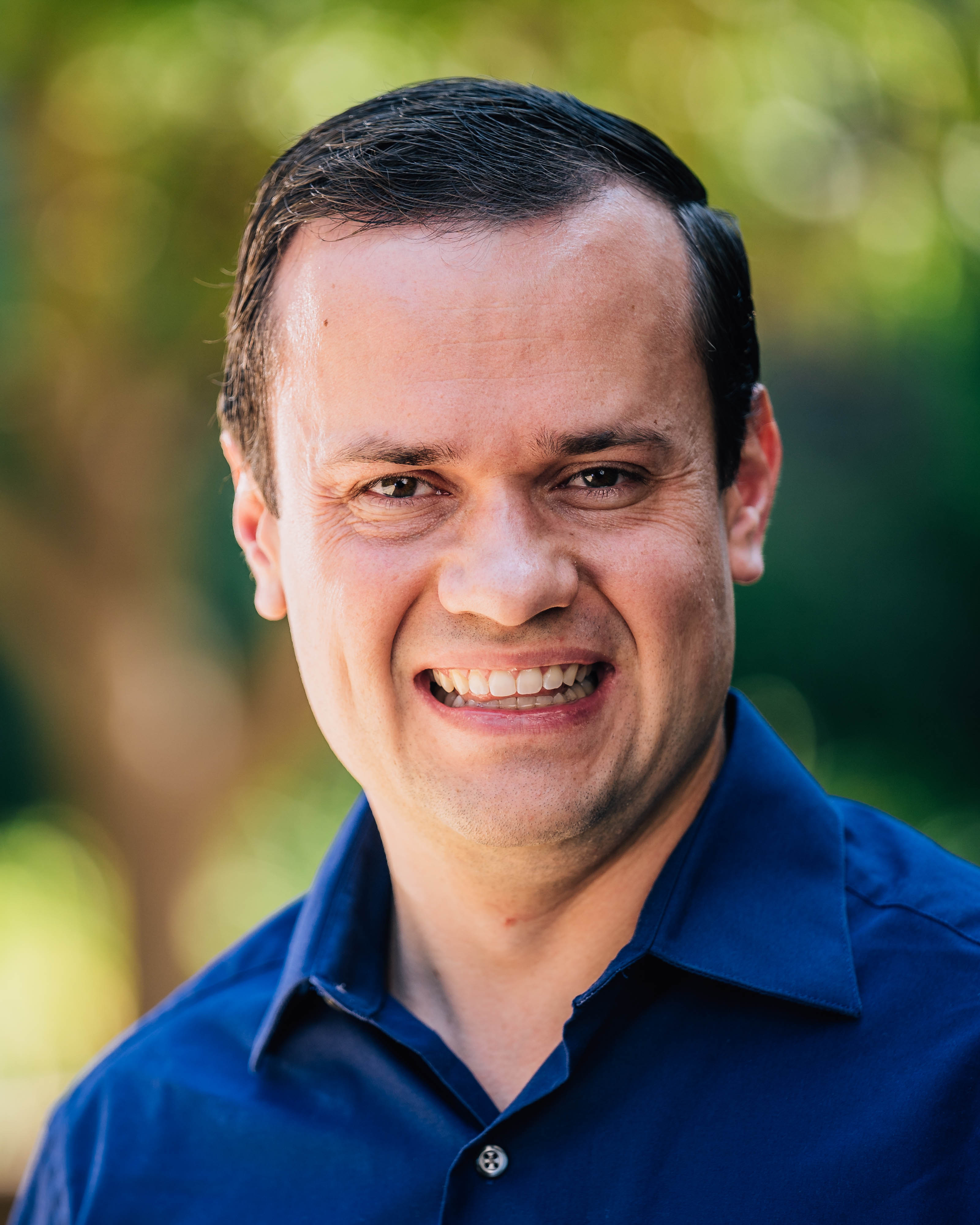}}]{Kevin R. Moon} is an assistant professor in the Department of Mathematics and Statistics at Utah State University (USU). He holds a B.S. and M.S. degree in Electrical Engineering from Brigham Young University and an M.S. degree in Mathematics and a Ph.D. in Electrical Engineering from the University of Michigan. Prior to joining USU in 2018, he was a postdoctoral scholar (2016-2018) in the Genetics Department and the Applied Mathematics Program at Yale University. His research interests are in the development of theory and applications in machine learning, information theory, deep learning, and manifold learning. More specifically, he is interested in dimensionality reduction, data visualization, ensemble methods, nonparametric estimation, information theoretic functionals, neural networks, and applications in biomedical data, finance, engineering, and ecology.
\end{IEEEbiography}

\clearpage
\appendices

\onecolumn
\counterwithin{figure}{section}
\counterwithin{table}{section}

\section{Additional Experimental Results}
\label{sec:appendix}
Here we present additional experimental results using RF-GAP. 


\subsection{Empirical Validation of Theorems~\ref{thm:regression} and~\ref{thm:classification}}

Here we present the full results for the experiments described in Section~\ref{sec:results}. The RF-GAP error rates match those of the RF.


\begin{table}[ht]
\centering
\caption{The random forest test average MSE (for continuously labeled data) and error rates (for categorically labeled data) of the random forest are presented in the RF column. The other columns represent the differences between the random forest and proximity-weighted predictions for regression and the proportion of mismatched predictions for classification. The RF-GAP proximity predictions nearly perfectly match the random forest predictions. Each experiment was repeated five times. The same random forest was used to construct each proximity for each random seed. RFProxIH cannot be used when categorical variables are present or for regression problems. Standard errors are in parentheses.} 
\label{tab:test_differences}
\begin{tabular}{|c||cc||cc|cc|cc|cc|cr|}
  \hline
Dataset & RF &  & RF-GAP &  & Original &  & OOB &  & PBK &  & RFProxIH &  \\ 
  \hline
Arrhythmia & 0.039 & (0.07) & \textbf{0.004} & (0.00) & 0.084 & (0.01) & 0.069 & (0.01) & 0.079 & (0.01) & & (NA)\\ 
   \hline
Auto-Mpg & 0.004 & (0.00) & \textbf{0.001} & (0.00) & 0.005 & (0.00) & 0.009 & (0.00) & 0.007 & (0.00) & & (NA) \\ 
   \hline
Balance Scale & 0.156 & (0.01) & \textbf{0.004} & (0.00) & 0.061 & (0.02) & 0.061 & (0.02) & 0.062 & (0.02) & 0.070 & (0.02) \\ 
   \hline
Banknote & 0.008 & (0.01) & \textbf{0} & (0.00) & 0.012 & (0.01) & 0.013 & (0.01) & 0.029 & (0.01) & 0.009 & (0.00) \\ 
   \hline
Breast Cancer & 0.033 & (0.00) & \textbf{0} & (0.00) & 0.016 & (0.01) & 0.016 & (0.01) & 0.018 & (0.00) & 0.013 & (0.01) \\ 
   \hline
Car & 0.043 & (0.01) &\textbf{ 0.011} & (0.00) & 0.028 & (0.01) & 0.033 & (0.01) & 0.080 & (0.02) &  & (NA) \\ 
   \hline
Diabetes & 0.236 & (0.02) & \textbf{0.001} & (0.00) & 0.052 & (0.02) & 0.072 & (0.01) & 0.057 & (0.01) & 0.064 & (0.01) \\ 
   \hline
E. Coli & 0.129 & (0.03) & \textbf{0.004} & (0.01) & 0.065 & (0.03) & 0.083 & (0.02) & 0.077 & (0.02) & 0.057 & (0.00) \\ 
   \hline
Glass & 0.228 & (0.05) & \textbf{0} & (0.00) & 0.094 & (0.03) & 0.103 & (0.03) & 0.109 & (0.02) & 0.112 & (0.02) \\ 
   \hline
Heart Disease & 0.435 & (0.05) & \textbf{0} & (0.00) & 0.101 & (0.04) & 0.136 & (0.03) & 0.116 & (0.05) &  & (NA) \\ 
   \hline
Hill Valley & 0.444 & (0.02) & \textbf{0} & (0.00) & 0.134 & (0.04) & 0.230 & (0.04) & 0.155 & (0.04) & 0.126 & (0.02) \\ 
   \hline
Ionosphere & 0.065 & (0.01) & \textbf{0.002} & (0.00) & 0.051 & (0.01) & 0.053 & (0.02) & 0.053 & (0.01) & 0.036 & (0.01) \\ 
   \hline
Iris & 0.031 & (0.01) &\textbf{ 0} & (0.00) & 0.004 & (0.01) & 0.009 & (0.01) & 0.013 & (0.01) & 0.013 & (0.01) \\ 
   \hline
Liver & 0.284 & (0.03) & \textbf{0.003 }& (0.00) & 0.098 & (0.03) & 0.153 & (0.04) & 0.106 & (0.04) &  & (NA) \\ 
   \hline
Lymphography & 0.164 & (0.05) & \textbf{0.005} & (0.01) & 0.050 & (0.04) & 0.064 & (0.05) & 0.068 & (0.06) & 0.041 & (0.02) \\ 
   \hline
Optical Digits & 0.024 & (0.01) & \textbf{0} & (0.00) & 0.020 & (0.00) & 0.021 & (0.00) & 0.023 & (0.00) & 0.012 & (0.00) \\ 
   \hline
Parkinsons & 0.078 & (0.04) & \textbf{0} & (0.00) & 0.068 & (0.02) & 0.078 & (0.02) & 0.075 & (0.02) & 0.047 & (0.02) \\ 
   \hline
RNASeq & 0.005 & (0.00) &\textbf{0} & (0.00) & 0.002 & (0.00) & 0.003 & (0.00) & 0.003 & (0.00) & 0.002 & (0.00) \\ 
   \hline
Seeds & 0.057 & (0.03) & \textbf{0} & (0.00) & 0.027 & (0.01) & 0.027 & (0.01) & 0.030 & (0.02) & 0.017 & (0.01) \\ 
   \hline
Sonar & 0.174 & (0.04) &\textbf{ 0.003 }& (0.01) & 0.071 & (0.04) & 0.097 & (0.02) & 0.081 & (0.04) & 0.077 & (0.04) \\ 
   \hline
Tic-Tac-Toe & 0.063 & (0.02) &\textbf{ 0.002} & (0.00) & 0.059 & (0.01) & 0.059 & (0.01) & 0.072 & (0.01) &  & (NA) \\ 
   \hline
Titanic & 0.185 & (0.03) & \textbf{0.010} & (0.01) & 0.049 & (0.01) & 0.055 & (0.01) & 0.057 & (0.01) &  & (NA) \\ 
   \hline
Waveform & 0.143 & (0.01) & \textbf{0.001} & (0.00) & 0.030 & (0.00) & 0.035 & (0.00) & 0.030 & (0.00) & 0.047 & (0.01) \\ 
   \hline
Wine & 0.023 & (0.02) & \textbf{0} & (0.00) & 0.011 & (0.02) & 0.011 & (0.02) & 0.019 & (0.02) & 0.011 & (0.02) \\ 
   \hline
\end{tabular}
\end{table}


\begin{table}[ht]
\centering
\caption{This table represents the differences in proximity-weighted predictions using the training set. These predictions were compared with the random forest OOB predictions. RF-GAP predictions most reliably match the random forest predictions. } 
\label{tab:training_differences}
\begin{tabular}{|c||cc||cc|cc|cc|cc|cr|}
  \hline
Dataset & RF &  & RF-GAP &  & Original &  & OOB &  & PBK &  & RFProxIH &  \\ 
  \hline
Arrhythmia & 0.064 & (0.00) & \textbf{0.006} & (0.00) & 0.078 & (0.00) & 0.066 & (0.00) & 0.074 & (0.00) &  & (NA) \\ 
   \hline
Auto-Mpg & 0.005 & (0.00) & \textbf{0.001} & (0.00) & 0.021 & (0.00) & 0.012 & (0.00) & 0.017 & (0.00) & & (NA) \\ 
   \hline
Balance Scale & 0.150 & (0.02) &\textbf{ 0.010} & (0.01) & 0.077 & (0.01) & 0.069 & (0.01) & 0.079 & (0.01) & 0.080 & (0.01) \\ 
   \hline
Banknote & 0.006 & (0.00) & \textbf{0 }& (0.00) & 0.004 & (0.00) & 0.014 & (0.00) & 0.019 & (0.01) & 0.003 & (0.00) \\ 
   \hline
Breast Cancer & 0.033 & (0.00) & \textbf{0 }& (0.00) & 0.015 & (0.00) & 0.021 & (0.00) & 0.018 & (0.00) & 0.011 & (0.00) \\ 
   \hline
Car & 0.043 & (0.01) & \textbf{0.010} & (0.00) & 0.023 & (0.00) & 0.022 & (0.01) & 0.058 & (0.01) &  & (NA) \\ 
   \hline
Diabetes & 0.243 & (0.01) & \textbf{0.002} & (0.00) & 0.151 & (0.01) & 0.081 & (0.02) & 0.129 & (0.01) & 0.151 & (0.01) \\ 
   \hline
E. Coli & 0.140 & (0.02) & \textbf{0.003} & (0.00) & 0.071 & (0.01) & 0.071 & (0.02) & 0.076 & (0.01) & 0.078 & (0.01) \\ 
   \hline
Glass & 0.215 & (0.03) & \textbf{0.003} & (0.00) & 0.137 & (0.02) & 0.119 & (0.02) & 0.123 & (0.03) & 0.141 & (0.03) \\ 
   \hline
Heart Disease & 0.417 & (0.02) & \textbf{0.008} & (0.00) & 0.301 & (0.02) & 0.221 & (0.01) & 0.270 & (0.02) &  & (NA) \\ 
   \hline
Hill Valley & 0.419 & (0.01) & \textbf{0.001} & (0.00) & 0.331 & (0.02) & 0.213 & (0.02) & 0.298 & (0.02) & 0.317 & (0.02) \\ 
   \hline
Ionosphere & 0.070 & (0.01) & \textbf{0} & (0.00) & 0.030 & (0.00) & 0.054 & (0.01) & 0.030 & (0.00) & 0.029 & (0.00) \\ 
   \hline
Iris & 0.065 & (0.01) & \textbf{0} & (0.00) & 0.038 & (0.02) & 0.010 & (0.01) & 0.034 & (0.01) & 0.044 & (0.01) \\ 
   \hline
Liver & 0.296 & (0.02) & \textbf{0.001} & (0.00) & 0.224 & (0.03) & 0.149 & (0.03) & 0.200 & (0.03) &  & (NA) \\ 
   \hline
Lymphography & 0.160 & (0.02) & \textbf{0.004} & (0.01) & 0.113 & (0.02) & 0.062 & (0.02) & 0.092 & (0.02) & 0.096 & (0.01) \\ 
   \hline
Optical Digits & 0.020 & (0.00) & \textbf{0} & (0.00) & 0.011 & (0.00) & 0.020 & (0.00) & 0.011 & (0.00) & 0.012 & (0.00) \\ 
   \hline
Parkinsons & 0.094 & (0.01) &\textbf{ 0.003 }& (0.00) & 0.040 & (0.01) & 0.078 & (0.01) & 0.065 & (0.03) & 0.040 & (0.01) \\ 
   \hline
RNAseq & 0.003 & (0.00) &\textbf{ 0} & (0.00) & 0.003 & (0.00) & 0.005 & (0.00) & 0.002 & (0.00) & 0.003 & (0.00) \\ 
   \hline
Seeds & 0.078 & (0.01) & \textbf{0.001} & (0.00) & 0.045 & (0.01) & 0.027 & (0.01) & 0.043 & (0.01) & 0.053 & (0.01) \\ 
   \hline
Sonar & 0.184 & (0.02) & \textbf{0.004 }& (0.01) & 0.163 & (0.03) & 0.104 & (0.03) & 0.156 & (0.03) & 0.171 & (0.03) \\ 
   \hline
Tic-Tac-Toe & 0.067 & (0.00) & \textbf{0.004} & (0.00) & 0.047 & (0.00) & 0.032 & (0.01) & 0.024 & (0.00) &  & (NA) \\ 
   \hline
Titanic & 0.182 & (0.01) & \textbf{0.011} & (0.01) & 0.076 & (0.01) & 0.070 & (0.01) & 0.074 & (0.01) &  & (NA) \\ 
   \hline
Waveform & 0.145 & (0.00) & \textbf{0.002} & (0.00) & 0.104 & (0.00) & 0.038 & (0.01) & 0.099 & (0.00) & 0.117 & (0.00) \\ 
   \hline
Wine & 0.019 & (0.01) & \textbf{0} & (0.00) & 0.019 & (0.01) & 0.006 & (0.00) & 0.018 & (0.01) & 0.019 & (0.01) \\ 
   \hline
\end{tabular}
\end{table}



\subsection{Proximity-Weighted Predictions and Sample Size}
\label{sub:sample_size}

\begin{figure}[bh!]
    \centering
    \includegraphics[width = \textwidth]{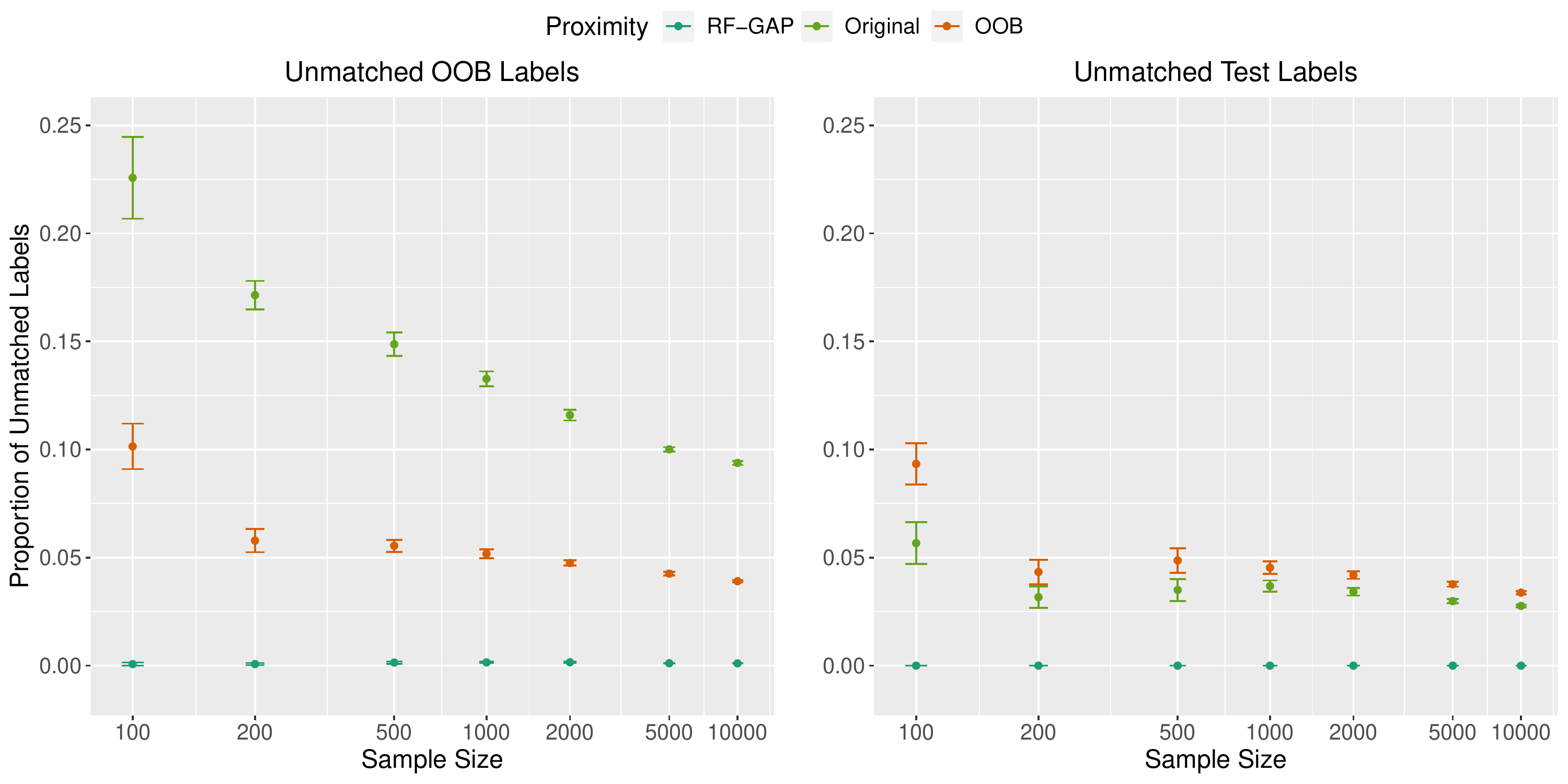}
    \caption{In this figure, we assess the effects of sample size on the proportion of proximity-weighted predictions which match the random forest's predictions on a synthetic dataset. The experiment was repeated 20 times. On the left, proximity-weighted training predictions are compared with the random forest's OOB predictions.   The means and standard errors are presented in the plot. While the Original and OOB proximity predictions match the random forest predictions as the same size increases, the difference does not appear to be approaching zero. In contrast,  RF-GAP matches the random forest predictions (up to tie breaks) regardless of the sample size.}
    \label{fig:sample_size}
\end{figure}

\begin{figure}[htb!]
    \centering
    \includegraphics[width = .45\textwidth]{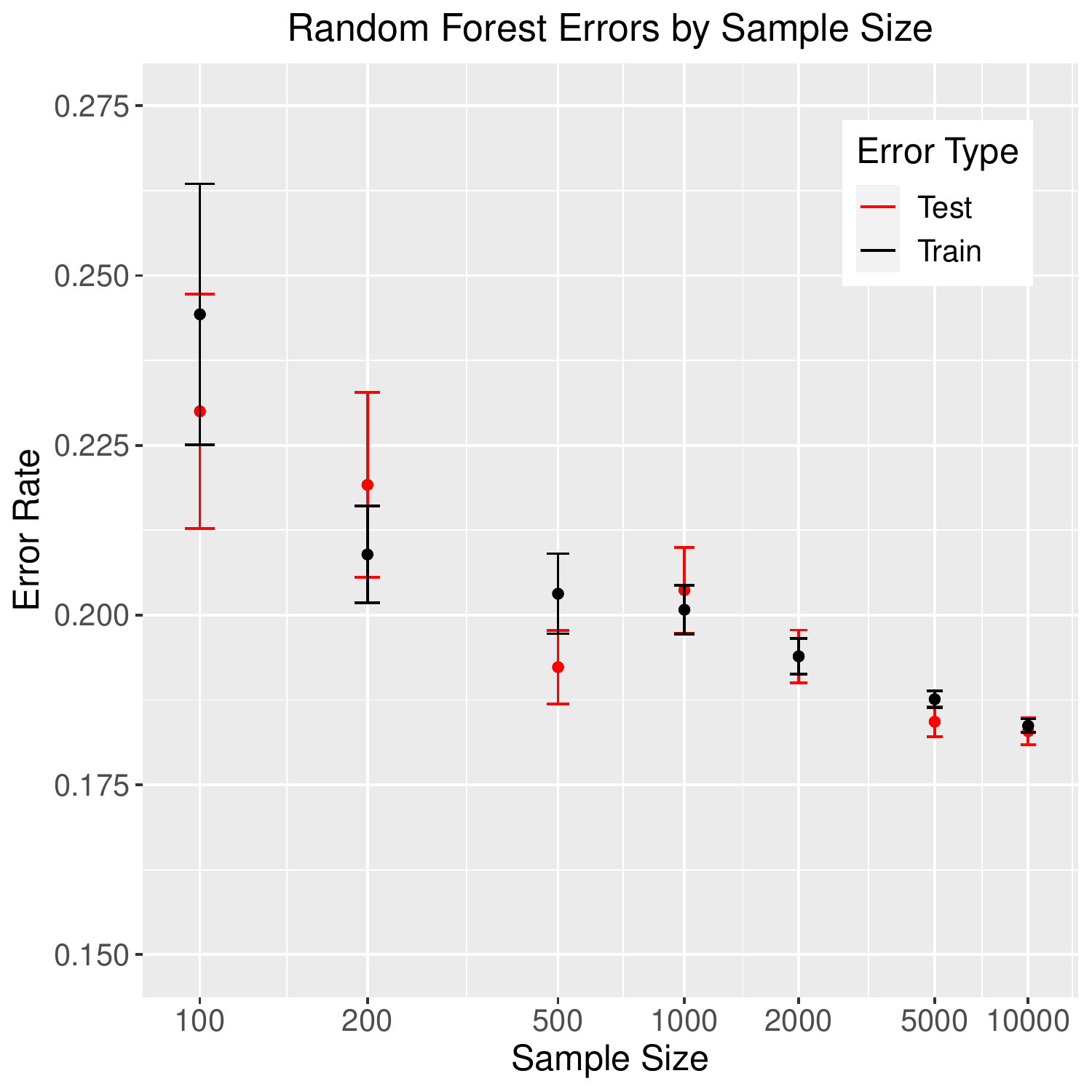}
    \caption{The random forest OOB and Test errors using the simulated datasets used in Figure~\ref{fig:sample_size}. As expected, both error rates tend to decrease with the sample size. This decrease may partially explain the higher proportion of proximity-weighted prediction matches for the OOB and Original proximities in Figure~\ref{fig:sample_size}.}
    \label{fig:sample_errors}
\end{figure}

Here we assess the effects of sample size on the random forest proximity-weighted predictions. Due to the computational complexity of PBK and RFProxIH, we only compared the Original, OOB, and RF-GAP proximities. For this experiment, we generated synthetic datasets with  binary class labels and sample sizes of 100, 200, 500, 1000, 2000, 5000, and 10000. Each dataset had ten normally-distributed variables. For class 0, each of the ten variables followed a standard normal distribution. For class 1, the means of the normally distributed variables were linearly spaced between 0 and 1, each with a variance of 1. This setup provides a range of variables with varying importance for determining class. The experiment was repeated 20 times. 

The results are shown in Figure~\ref{fig:sample_size}. The overall random forest prediction errors are found in Figure~\ref{fig:sample_errors}. The Original and OOB proximity predictions more closely match the random forest predictions as the sample size increases, but this difference does not appear to tend toward zero. It is interesting to note that the original proximities are better approximations for the random forest test predictions than the OOB proximities, although the reverse is true when matching the RF OOB predictions. The OOB proximity test and training predictions both appear to be converging to the same proportion of unmatched random forest predictions for the sample sizes considered. In contrast,  the RF-GAP predictions match the random forest predictions (up to tie breaks) regardless of the sample size.


\subsection{RF-GAP and Symmetry}

\begin{figure}[htb!]
    \centering
    \includegraphics[width =.45\textwidth]{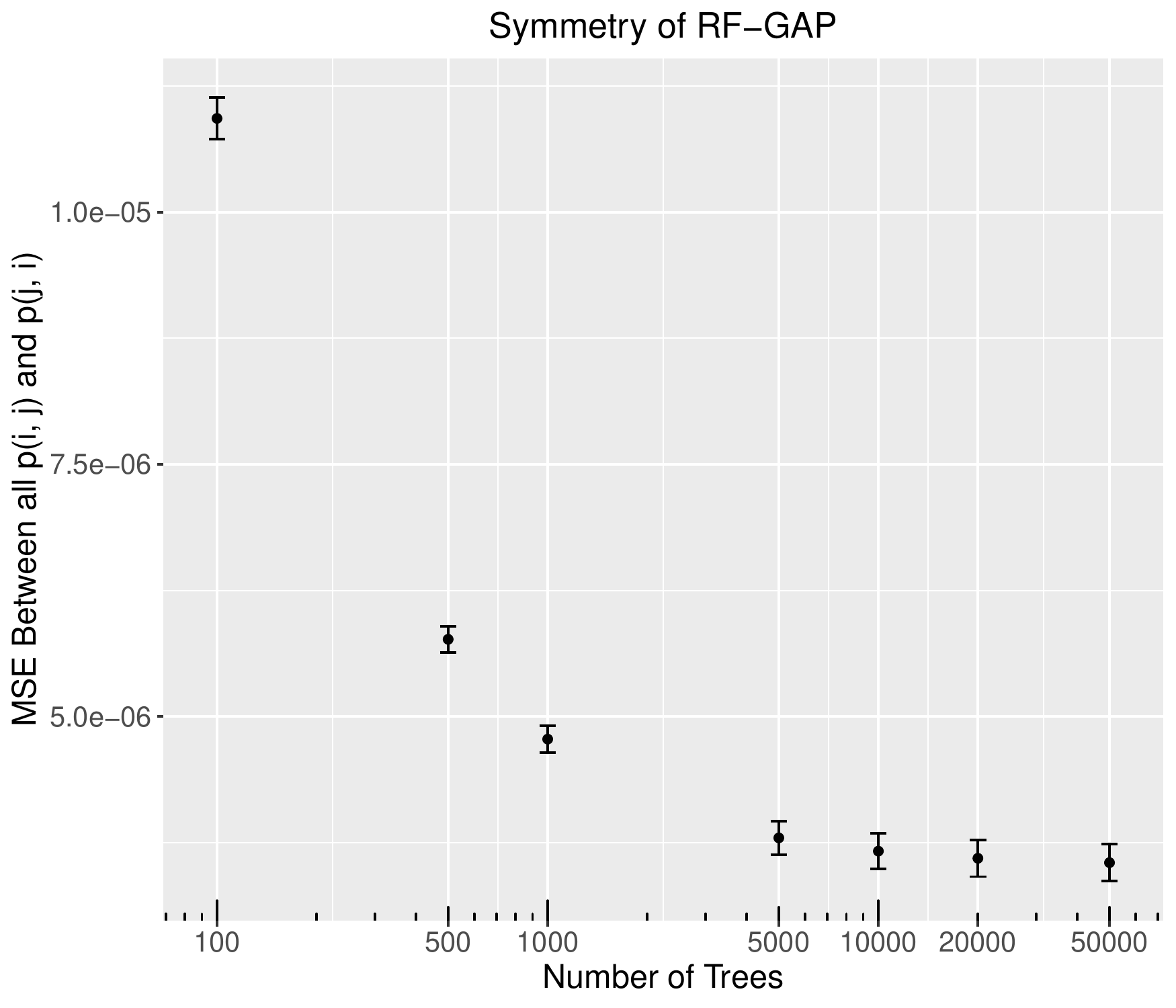}
    \caption{Here we show that RF-GAP proximities become more symmetric as the number of trees increases. The data used here had 500 observations with binary class labels and 10 normally-distributed features, as described in Section~\ref{sub:sample_size}.  The MSE between the proximity matrix and its transpose was computed across each run. The mean and standard deviations are provided in the scatter plot.}
    \label{fig:symmetry}
\end{figure}

Although the RF-GAP proximities are not symmetric, we show that their symmetry increases with the number of trees in the forest. We used the same synthetic dataset described in Section~\ref{sub:sample_size} using 500 observations. The experiment was replicated 10 times. The results are shown in Figure~\ref{fig:symmetry}.

\subsection{Minimum Node Size}
\begin{figure}[htb!]
    \centering
    \includegraphics[width = \textwidth]{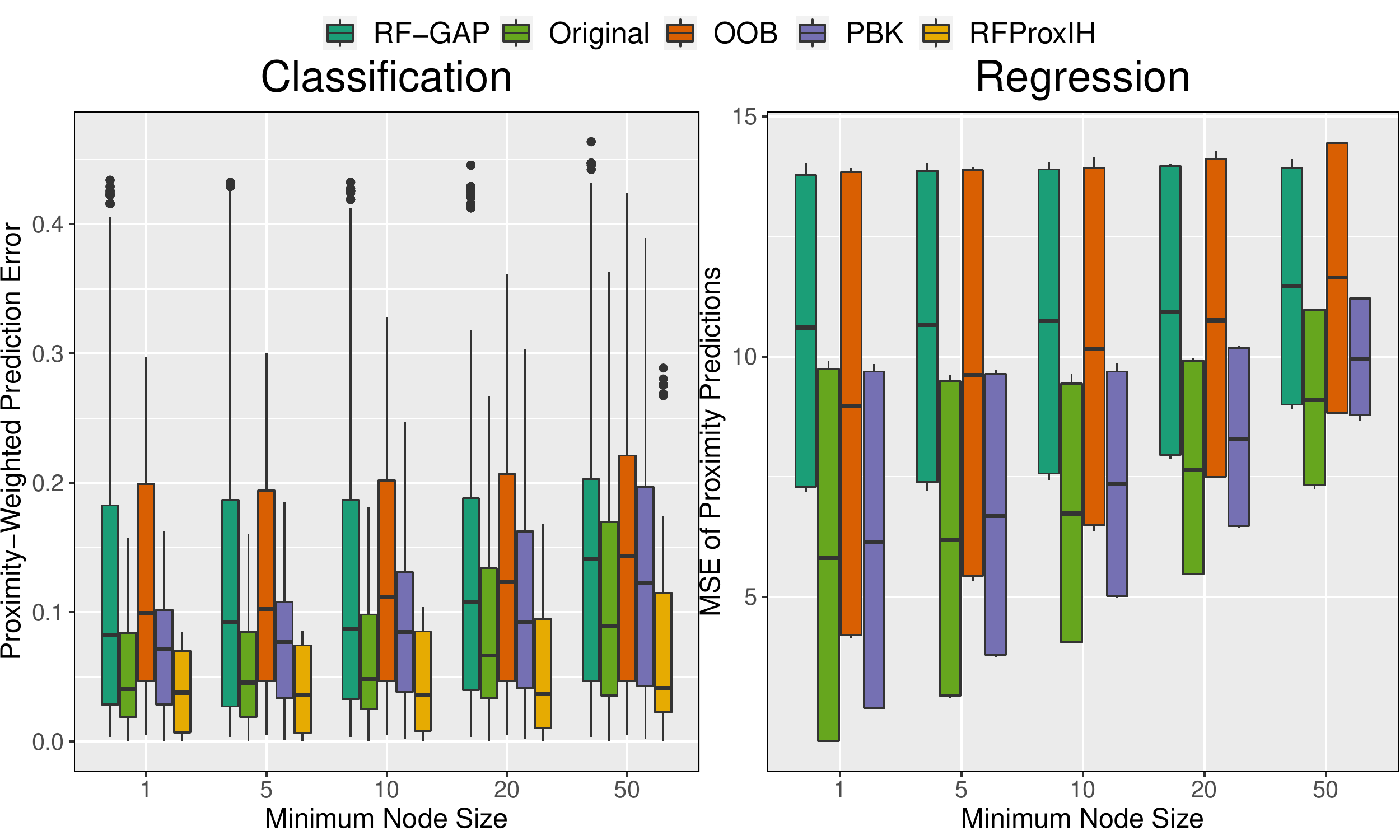}
    \caption{Here we display the distributions of errors corresponding to the proximity-weighted training predictions across the minimum node sizes 1, 5, 10, 20, and 50. Predictions were made using five random seeds and each of the datasets listed in Table~\ref{tab:datasets}. The errors presented are proximity-weighted training errors. Each of the errors tends to increase as the minimum node size increases.}
    \label{fig:node_size_error}
\end{figure}

We show in Figure~\ref{fig:node_size_error} that both the random forest and proximity-weighted training errors tend to increase with the minimum node size. See Section~\ref{sec:proofs} for further discussion.

\subsection{Number of Trees}

\begin{figure}
    \centering
    \includegraphics[width=\textwidth]{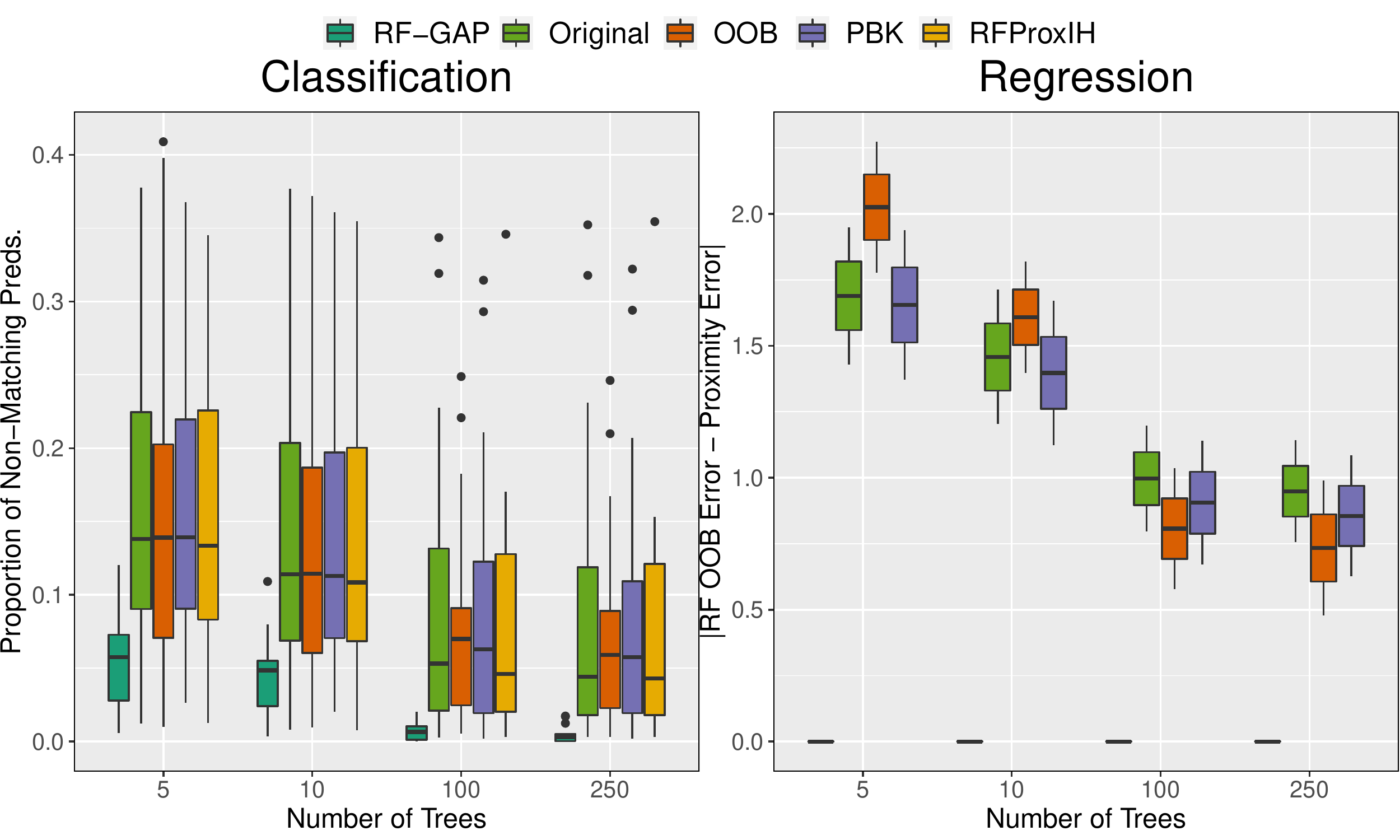}
    \caption{Proximity-weighted prediction matches using varying numbers of trees. Generally, RF-GAP produces predictions that match those of the random forest except for tie-break cases. For regression, the matches are exact. Each proximity definition tends to match better with an increased number of trees.}
    \label{fig:num_trees}
\end{figure}

We show to what extent proximity-weighted predictions match the random forest predictions when considering different sizes of forests. It has been shown that larger forests produce more accurate predictions~\cite{Breiman2001randomforests, probst2017tuneornot}, but there is a consideration for computational complexity.  For all other experiments in this paper, we used the default value of 500 trees to train each forest.  Here, we show that the RF-GAP predictions still match the random forest predictions with small numbers of trees (5, 10, 50, 100, 250). Figure~\ref{fig:num_trees} shows the results compiled across the same 24 datasets described in Table~\ref{tab:datasets}. In the regression context, the matches are exact. In the classification context, we see a decrease in matches as the number of trees decreases. This is due to an increased need for tie-breaking, which is not needed for regression.  

\subsection{Agreement Between Proximities}

\begin{figure}[htb!]
    \centering
    \includegraphics[width = .49\textwidth]{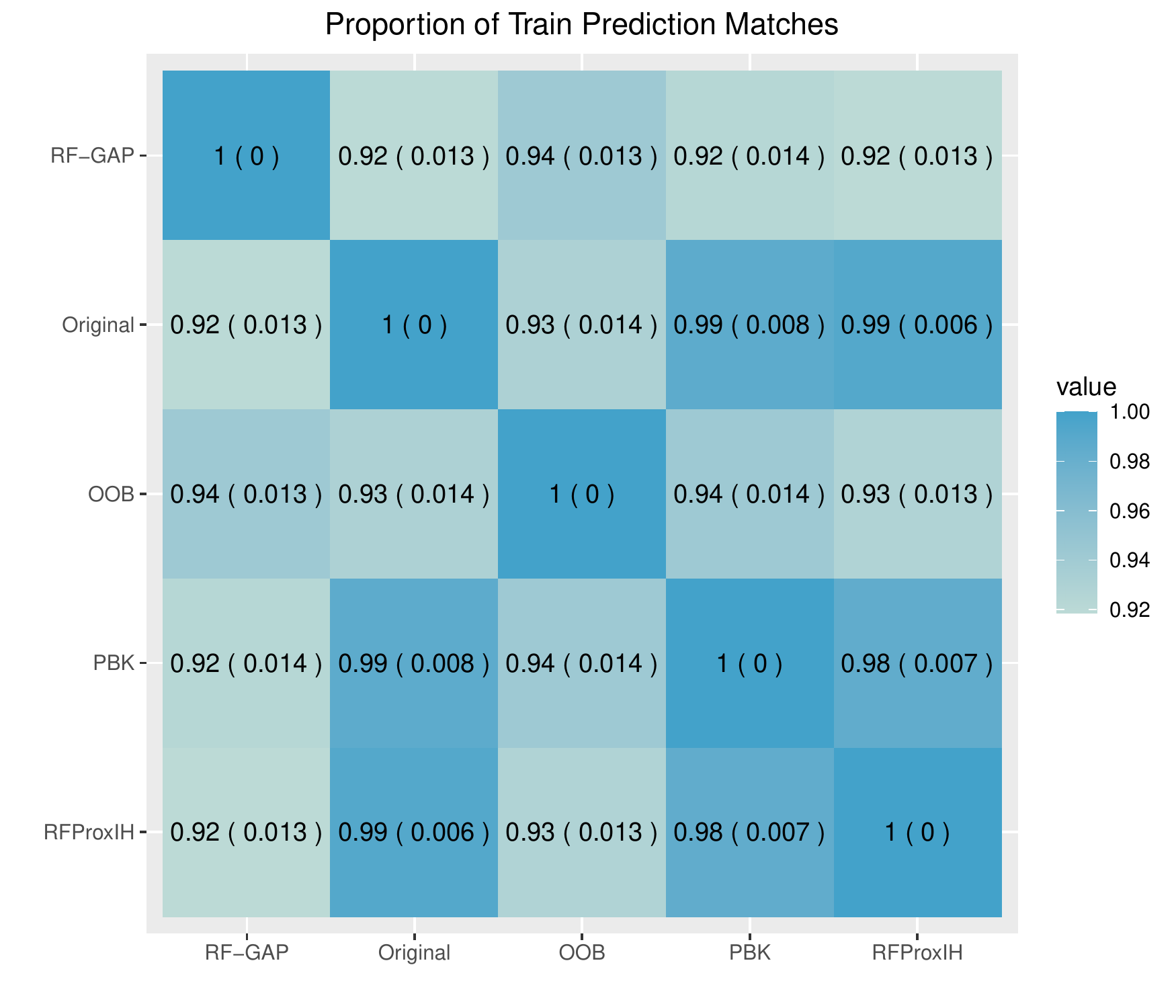}
    \includegraphics[width = .49\textwidth]{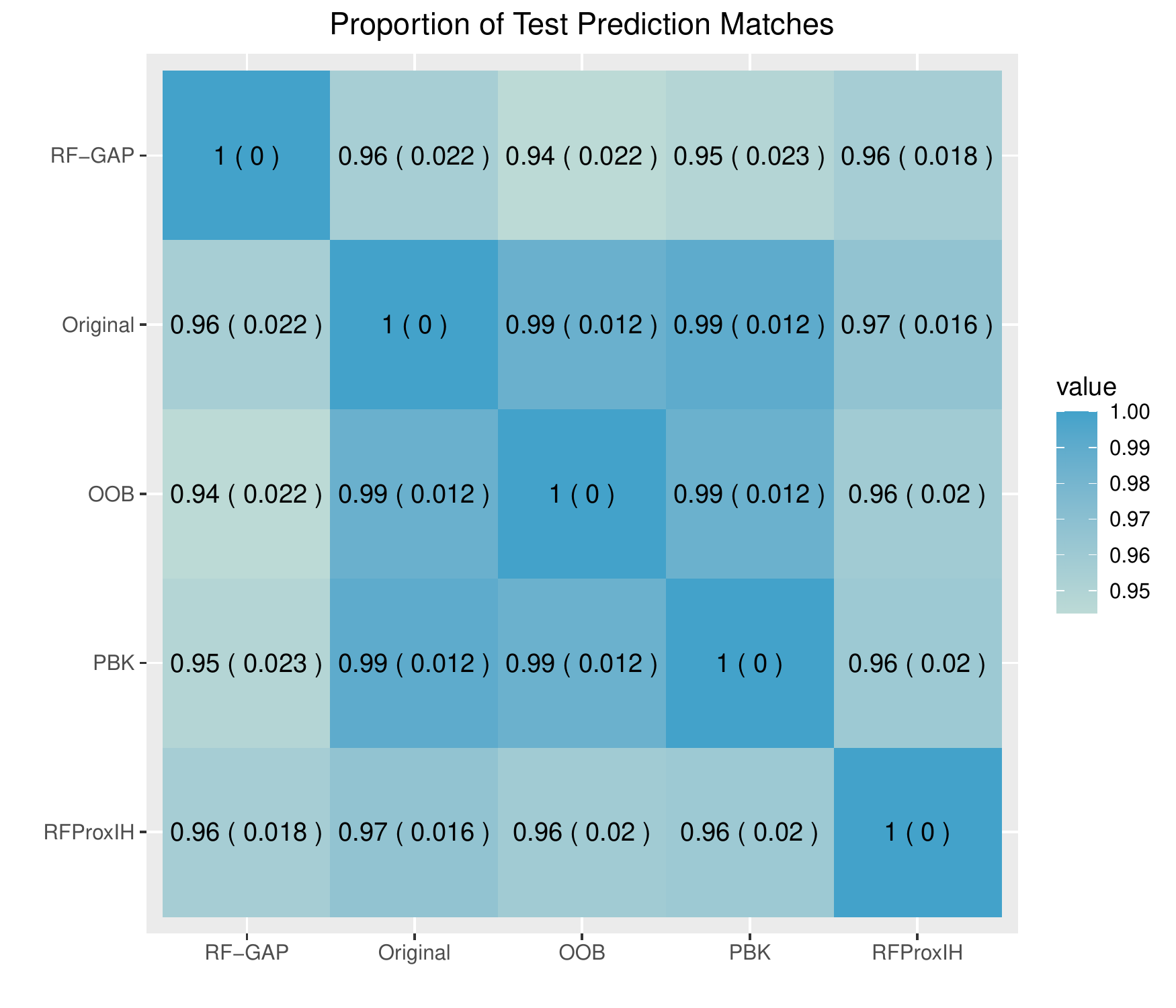}
    \caption{The proportions of proximity-weighted prediction agreements for the train (left) and test (right) sets across categorical-labeled datasets as found in Table~\ref{tab:datasets}. RF-GAP agrees less often with the others, while the original, PBK, and RFProxIH tend to agree most often for the training data. The original, OOB, and PBK proximities tend to agree the most often for the test data.}
    \label{fig:test_agreement}
\end{figure}

In addition to comparing proximity-weighted predictions with the random forest predictions, we also compared the level of agreement from proximity to proximity. The proximity-weighted test predictions varied less across proximities, but the largest differences were found comparing RF-GAP to the others. The original, OOB, and PBK definitions had the highest level of agreement for the test predictions, as seen in Figure~\ref{fig:test_agreement}. Larger discrepancies were found when predicting training examples. In particular, RF-GAP proximities differed the most from each of the other proximities as their predictions match the random forest out-of-bag predictions while the others do not.

\subsection{Multidimensional Scaling and Outlier Detection}\label{subsec:mds_outlier}

\begin{figure*}
    \centering
    \includegraphics[width = \textwidth]{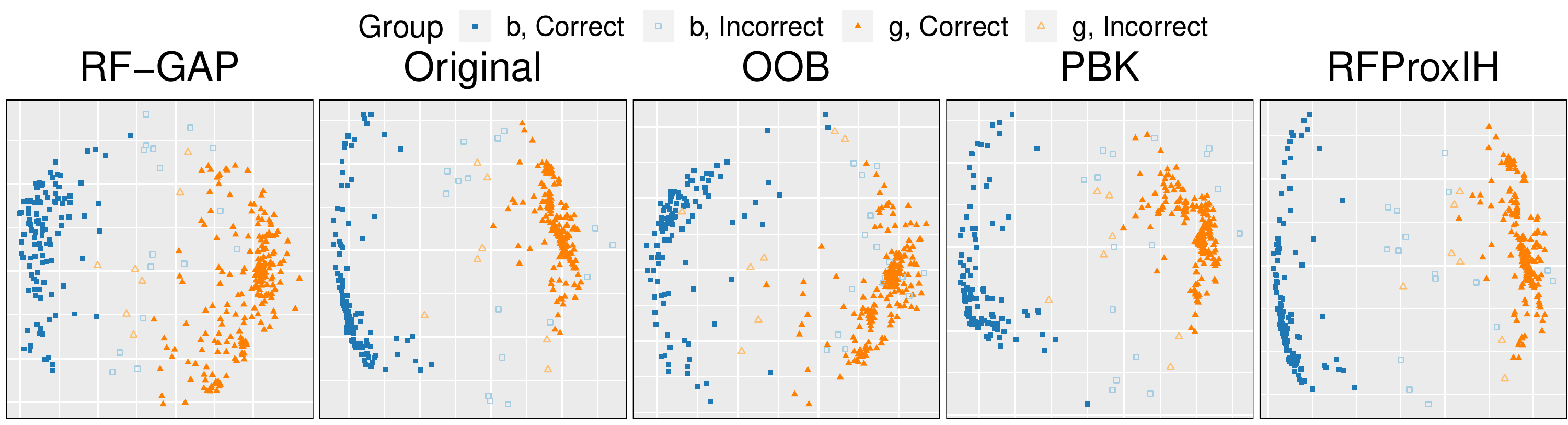}
    \caption{MDS applied to various random forest proximities on the Ionosphere dataset~\cite{UCI2019}. This binary classification problem predicts whether or not returned radar signals are representative of a structure (good) or not (bad). We see a similar pattern here regarding the MDS embeddings as in Figure~\ref{fig:sonar}. The class separations are somewhat exaggerated for the original, PBK, and RFProxIH proximities, while points clearly susceptible to misclassification are identifiable in the RF-GAP and OOB plots.}
    \label{fig:iono}
\end{figure*}

\begin{figure*}
    \centering
    \includegraphics[width = \textwidth]{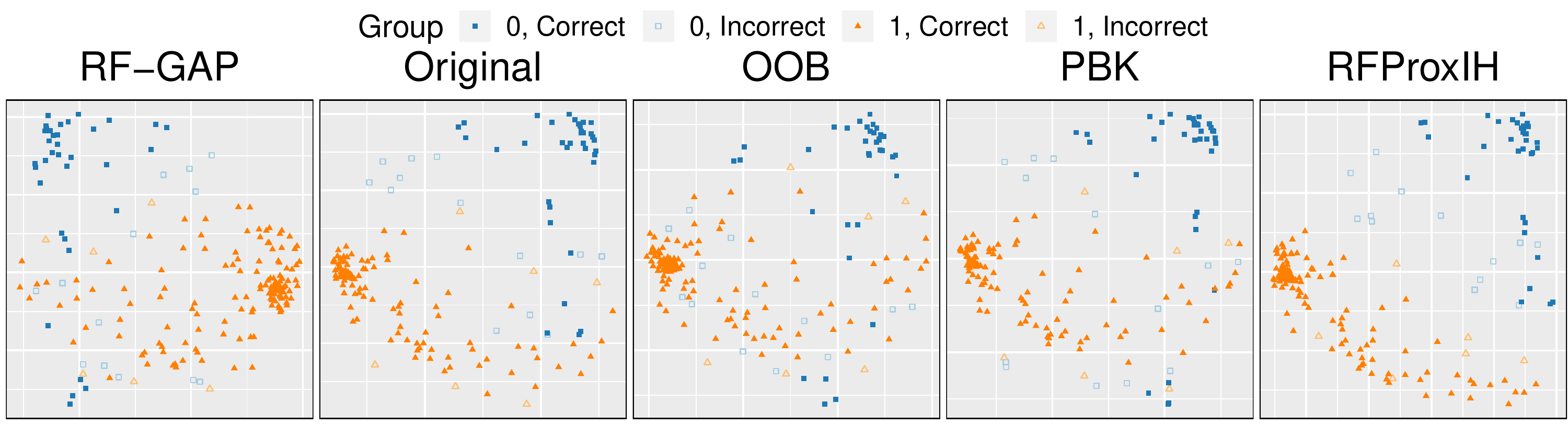}
    \caption{MDS applied to various random forest proximities on the Parkinson's dataset~\cite{UCI2019}, which tests whether machine learning algorithms can discriminate between healthy and unhealthy speech signals recorded from people with Parkinson's disease. From the RF-GAP embeddings, it is clear that misclassified points are on the borders or edges of the main clusters. This provides an example where random forest predictions correspond to proximity-weighted predictions.  This is not always clear in the other embeddings.  For example, the original MDS embeddings show a misclassified 1 (in the bottom right of the figure) which is surrounded by observations of the same class. RFProxIH shows a nearly perfectly linear separation between classes, which is unreasonable with a random forest error rate of 9.7\%.}
    \label{fig:park}
\end{figure*}

\begin{figure*}
    \centering
    \includegraphics[width = \textwidth]{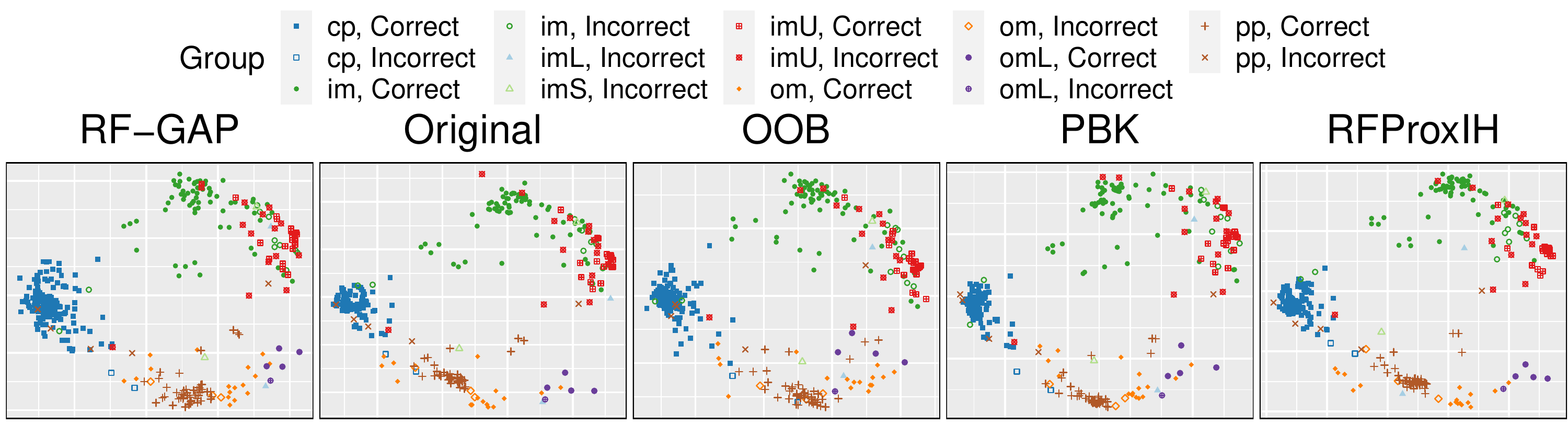}
    \caption{The E. coli dataset from the UCI repository with eight classes after applying MDS to the random forest proximities. RF-GAP and OOB show looser clusters compared with the others. This is suggestive of less overfitting of the training data.}
    \label{fig:ecoli}
\end{figure*}

\begin{figure*}
    \centering
    \includegraphics[width = \textwidth]{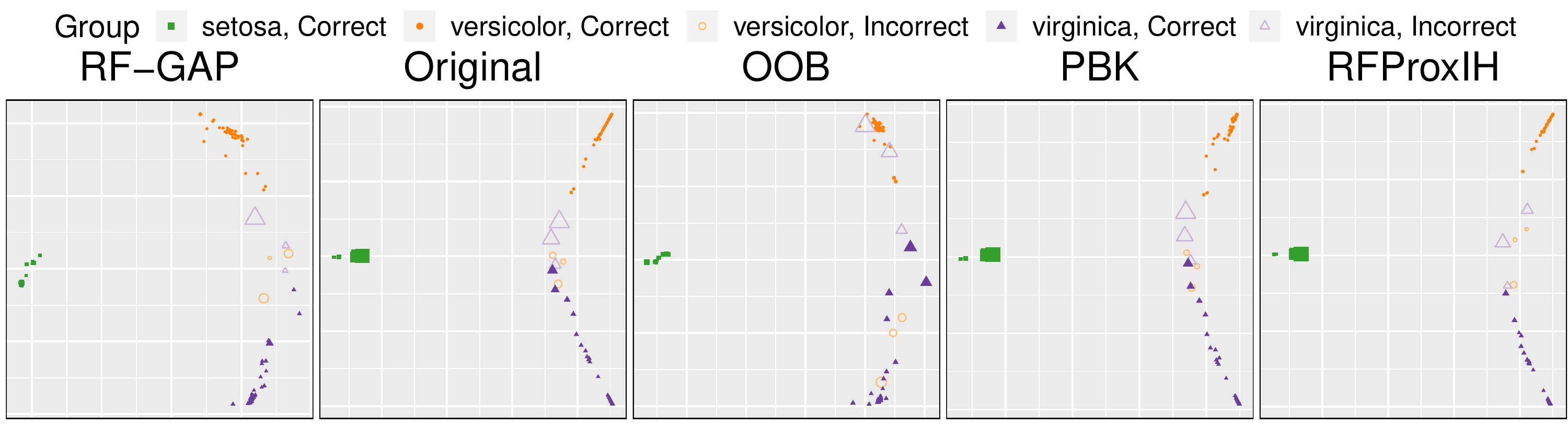}
    \caption{Fisher's Iris dataset~\cite{UCI2019} with MDS applied to the random forest proximities. Here, the point size is proportional to the outlier score provided by each method. In each case, observations with high outlier scores corresponded to misclassified points.}
    \label{fig:iris}
\end{figure*}

Here we provide additional examples of MDS applied to the various random forest proximities. In Figure~\ref{fig:iono}, we compare the plotted MDS embeddings on the Ionosphere data from the UCI~\cite{UCI2019}.  It is clear from the images that the random forest's misclassified points are typically found on the border between the two class clusters in the RF-GAP embeddings while this is not always the case for the other proximity measures. Additional figures (\ref{fig:park}, \ref{fig:ecoli}, and \ref{fig:iris}) are given to display MDS applied to the proximities on other datasets.  In Figure~\ref{fig:park} we see similar patterns which were displayed in Figure~\ref{fig:sonar}; exaggerated separation in the original and RFProxIH and excess noise in OOB.  RF-GAP seems to accurately portray why the misclassifications are made in the context of proximity-weighed predictions.  

\begin{figure*}
    \centering
    \includegraphics[width = \textwidth]{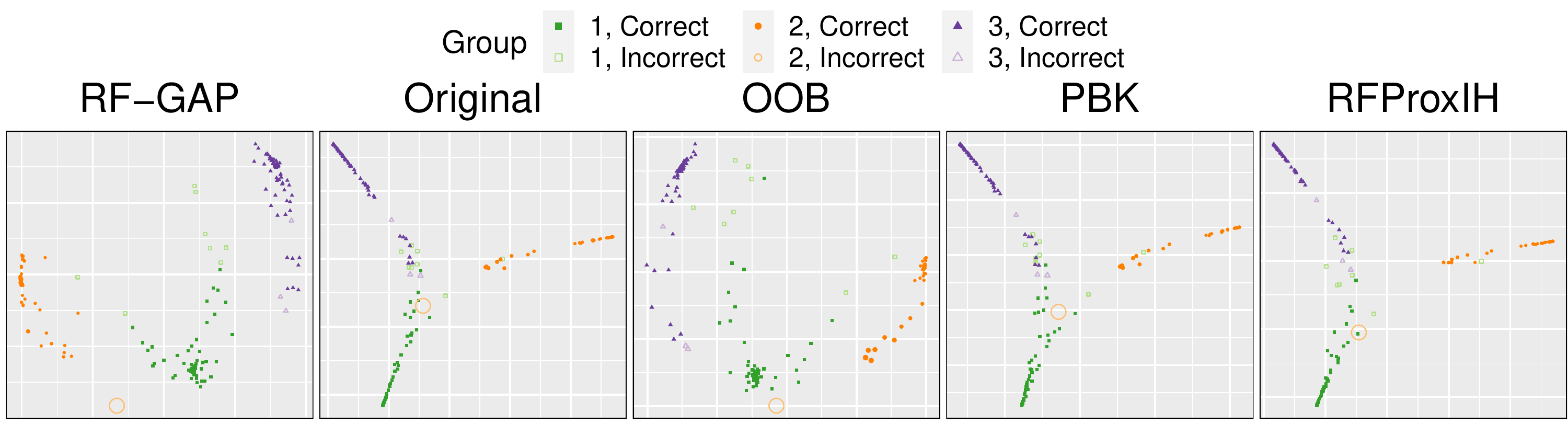}
    \caption{The seeds dataset compares three varieties of wheat seeds using geometric properties (e.g., width, length) as features. The OOB and RF-GAP proximities produce more cluster-like structures, vs. the branching seen by the other definitions. RF-GAP clearly shows why the misclassifications are taking place.}
    \label{fig:seeds}
\end{figure*}

\begin{figure*}
    \centering
    \includegraphics[width = \textwidth]{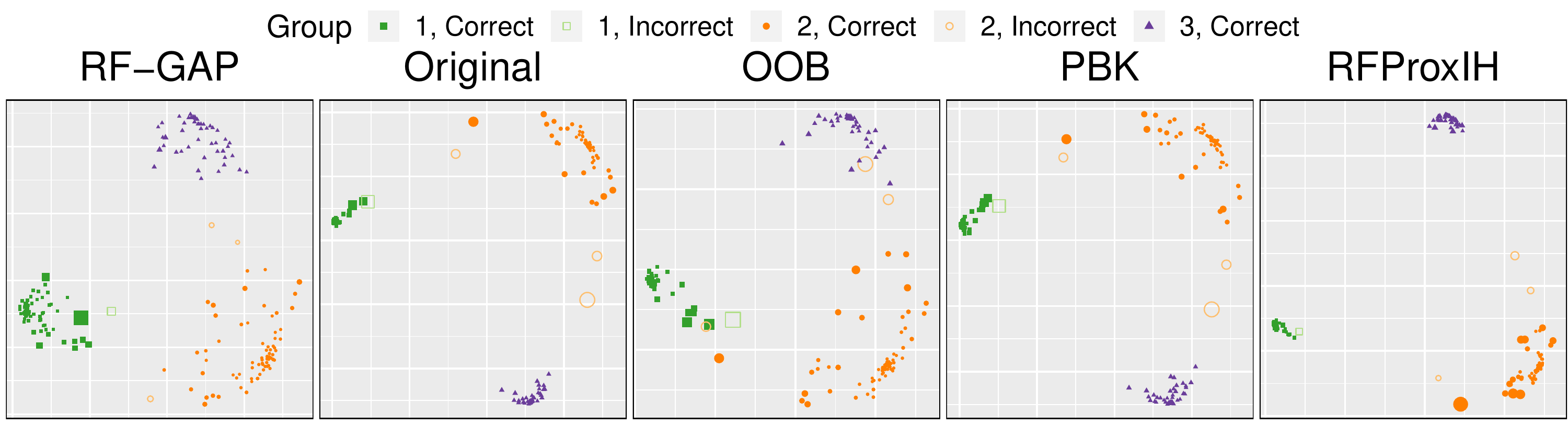}
    \caption{The wine dataset consists of three classes (corresponding to locations of cultivation) and 13 features. The tight clusters of each class using the original, PBK, and RFProxIH proximities suggests overfitting to the training data. It seems RF-GAP may show more sensitivity to outliers.}
    \label{fig:wine}
\end{figure*}

Figures~\ref{fig:iris}, \ref{fig:seeds}, and \ref{fig:wine} give additional examples of proximity-based outlier scores. Points that are farther from their respective class clusters can be viewed as outliers and are often misclassified. The point size is proportional to the outlier measure in the figure. This, however, may not be as clear when two or three points are far from their respective cluster but near to each other. 

\newpage

\subsection{Data Imputation}

\begin{table}
\centering
\scriptsize
\caption{Extended imputation results. For each dataset, values were removed completely at random in the amounts of 5\%, 10\%, 25\%, 50\%, and 75\%.   The two numbers directly below each dataset designate the number of observations and number of variables, respectively, for the listed dataset. The average  mean-squared errors (MSE) between the original and imputed values were recorded along with the standard deviations. RF-GAP generally outperforms the other definitions. Average rankings can be found in Table~\ref{tab:imp_ranks}.} 
\begin{tabular}{|l|c|cc|cc|cc|cc|cr|}
  \hline
Dataset & Proximity & 5\% &  & 10\% & & 25\% &  & 50\% &  & 75\% &  \\ 
  \hline
Arrhythmia &       RF-GAP & \textbf{0.637} & (0.05) & \textbf{0.913} & (0.06) & \textbf{1.463} & (0.06) & \textbf{2.142} & (0.06) & \textbf{2.783} & (0.06) \\ 
  452 &     Original & 0.665 & (0.05) & 0.950 & (0.06) & 1.512 & (0.06) & 2.188 & (0.06) & 2.792 & (0.05) \\ 
  279 &           OOB & 0.674 & (0.05) & 0.964 & (0.06) & 1.532 & (0.06) & 2.205 & (0.06) & 2.766 & (0.05) \\ 
   &  RFProxIH & NA & NA & NA & NA & NA & NA & NA & NA & NA & NA \\ 
   \hline
Auto-Mpg &     RF-GAP & 0.230 & (0.05) & 0.338 & (0.06) & 0.587 & (0.06) & 0.941 & (0.06) & 1.484 & (0.20) \\ 
  392 &   Original & \textbf{0.226} & (0.05) & \textbf{0.334} & (0.05) & \textbf{0.584} & (0.06) & \textbf{0.938} & (0.06) & \textbf{1.400} & (0.10) \\ 
  8 &         OOB & 0.232 & (0.05) & 0.342 & (0.06) & 0.595 & (0.06) & 0.955 & (0.06) & 1.425 & (0.10) \\ 
  & RFProxIH & NA & NA & NA & NA & NA & NA & NA& NA & NA & NA \\ 
   \hline
Balance Scale &      RF-GAP & \textbf{3.240} & (0.22) & \textbf{4.581 }& (0.22) & \textbf{7.288} & (0.23) & \textbf{10.278} & (0.24) & \textbf{12.553} & (0.26) \\ 
  645 &    Original & 3.287 & (0.22) & 4.649 & (0.22) & 7.385 & (0.23) & 10.336 & (0.24) & 12.561 & (0.27) \\ 
  4 &          OOB & 3.318 & (0.22) & 4.686 & (0.22) & 7.432 & (0.23) & 10.382 & (0.24) & 12.609 & (0.27) \\ 
   & RFProxIH & 3.305 & (0.23) & 4.674 & (0.23) & 7.414 & (0.23) & 10.351 & (0.25) & 12.571 & (0.27) \\ 
   \hline
Banknote &          RF-GAP & \textbf{1.372} & (0.12) & \textbf{1.965} & (0.12) & \textbf{3.211} & (0.13) & \textbf{4.798} & (0.12) & \textbf{ 6.176} & (0.12) \\ 
  1372 &        Original & 1.427 & (0.12) & 2.048 & (0.12) & 3.361 & (0.12) & 4.986 & (0.11) & 6.313 & (0.12) \\ 
  5 &              OOB & 1.443 & (0.12) & 2.065 & (0.12) & 3.379 & (0.12) & 4.994 & (0.12) & 6.319 & (0.12) \\ 
   &     RFProxIH & 1.420 & (0.12) & 2.042 & (0.12) & 3.357 & (0.12) & 4.985 & (0.12) & 6.314 & (0.12) \\ 
   \hline
Breast Cancer &      RF-GAP & \textbf{2.816} & (0.20) & \textbf{4.027} & (0.22) & \textbf{6.479} & (0.20) & \textbf{9.362} & (0.20) & \textbf{11.737} & (0.23) \\ 
  699 &    Original & 2.903 & (0.20) & 4.154 & (0.23) & 6.670 & (0.21) & 9.563 & (0.22) & 11.848 & (0.23) \\ 
  16 &          OOB & 2.891 & (0.20) & 4.125 & (0.23) & 6.604 & (0.21) & 9.483 & (0.21) & 11.800 & (0.23) \\ 
   & RFProxIH & 2.921 & (0.20) & 4.183 & (0.23) & 6.713 & (0.22) & 9.601 & (0.22) & 11.869 & (0.23) \\ 
   \hline
Diabetes &      RF-GAP & \textbf{0.710 }& (0.20) & \textbf{0.960} & (0.16) & \textbf{1.591} & (0.17) & \textbf{2.328} & (0.15) & \textbf{2.932} & (0.14) \\ 
  678 &    Original & 0.713 & (0.19) & 0.973 & (0.15) & 1.622 & (0.17) & 2.364 & (0.16) & 2.941 & (0.14) \\ 
  8 &          OOB & 0.713 & (0.19) & 0.971 & (0.15) & 1.616 & (0.17) & 2.360 & (0.15) & 2.941 & (0.14) \\ 
   & RFProxIH & 0.715 & (0.19) & 0.977 & (0.15) & 1.628 & (0.17) & 2.367 & (0.16) & 2.942 & (0.14) \\ 
   \hline
E. Coli &      RF-GAP & 0.862 & (0.14) & 1.235 & (0.12) & 1.991 & (0.11) & 2.873 & (0.13) & 3.673 & (0.14) \\ 
  336 &    Original & 0.886 & (0.14) & 1.267 & (0.12) & 2.031 & (0.12) & 2.909 & (0.14) & 3.706 & (0.17) \\ 
  8 &          OOB & 0.898 & (0.14) & 1.286 & (0.12) & 2.059 & (0.12) & 2.955 & (0.14) & 3.782 & (0.23) \\ 
   & RFProxIH & \textbf{0.831} & (0.23) & \textbf{1.156} & (0.36) & \textbf{1.774} & (0.67) & \textbf{2.170} & (1.26) & \textbf{2.146} & (1.80) \\ 
   \hline
Glass &      RF-GAP & \textbf{0.064} & (0.02) & \textbf{0.093} & (0.02) & \textbf{0.161} & (0.02) & 0.251 & (0.12) & 0.479 & (0.49) \\ 
  214 &    Original & 0.067 & (0.02) & 0.098 & (0.02) & 0.169 & (0.03) & 0.248 & (0.03) & 0.325 & (0.02) \\ 
  10 &          OOB & 0.068 & (0.02) & 0.099 & (0.02) & 0.169 & (0.03) &\textbf{ 0.246} & (0.03) & \textbf{0.323} & (0.02) \\ 
   & RFProxIH & 0.067 & (0.02) & 0.099 & (0.02) & 0.171 & (0.03) & 0.250 & (0.03) & 0.326 & (0.02) \\ 
   \hline
Heart Disease &      RF-GAP & \textbf{0.314} & (0.07) & \textbf{0.487} & (0.11) & \textbf{0.757} & (0.09) & \textbf{1.116} & (0.08) & 1.411 & (0.09) \\ 
  303 &    Original & 0.317 & (0.07) & 0.491 & (0.11) & 0.764 & (0.09) & 1.121 & (0.08) & 1.405 & (0.08) \\ 
  13 &          OOB & 0.316 & (0.07) & 0.489 & (0.11) & 0.761 & (0.09) & \textbf{1.116} & (0.08) & \textbf{1.399 }& (0.08) \\ 
   & RFProxIH & NA & NA & NA & NA & NA & NA & NA & NA & NA & NA \\ 
   \hline
Hill Valley &      RF-GAP & \textbf{1.418} & (0.20) & \textbf{4.093 }& (0.43) & \textbf{9.971} & (0.65) & \textbf{15.307} & (0.75) & \textbf{19.586} & (0.88) \\ 
  606 &    Original & 2.052 & (0.29) & 5.304 & (0.39) & 10.656 & (0.60) & 16.134 & (0.76) & 20.568 & (0.92) \\ 
  101 &          OOB & 1.673 & (0.26) & 4.996 & (0.42) & 10.526 & (0.59) & 15.984 & (0.75) & 20.489 & (0.92) \\ 
   & RFProxIH & 2.232 & (0.30) & 5.512 & (0.38) & 10.716 & (0.59) & 16.161 & (0.76) & 20.578 & (0.92) \\ 
   \hline
Ionosphere &      RF-GAP & \textbf{4.472} & (0.23) & \textbf{6.522} & (0.23) & \textbf{10.560} & (0.29) & \textbf{15.636} & (0.34) & \textbf{20.092} & (0.39) \\ 
  351 &    Original & 4.782 & (0.22) & 6.991 & (0.23) & 11.291 & (0.27) & 16.406 & (0.30) & 20.501 & (0.35) \\ 
  34 &          OOB & 4.745 & (0.22) & 6.933 & (0.23) & 11.184 & (0.27) & 16.299 & (0.31) & 20.446 & (0.35) \\ 
   & RFProxIH & 4.800 & (0.22) & 7.021 & (0.23) & 11.344 & (0.27) & 16.449 & (0.30) & 20.519 & (0.35) \\ 
   \hline
Iris &      RF-GAP & \textbf{0.211} & (0.04) & \textbf{0.300} & (0.04) & \textbf{0.494} & (0.04) & \textbf{0.716} & (0.04) & 0.935 & (0.05) \\ 
  150 &    Original & 0.219 & (0.04) & 0.310 & (0.04) & 0.509 & (0.04) & 0.727 & (0.04) & \textbf{0.934} & (0.05) \\ 
  4 &          OOB & 0.217 & (0.04) & 0.307 & (0.04) & 0.503 & (0.04) & 0.722 & (0.04) & 0.937 & (0.05) \\ 
   & RFProxIH & 0.221 & (0.04) & 0.312 & (0.05) & 0.512 & (0.04) & 0.730 & (0.04) & 0.936 & (0.05) \\ 
   \hline
Liver &      RF-GAP & \textbf{0.315} & (0.16) & \textbf{0.501} & (0.23) & \textbf{0.828} & (0.24) & \textbf{1.168} & (0.23) & \textbf{1.491} & (0.20) \\ 
  345 &    Original & 0.322 & (0.16) & 0.512 & (0.22) & 0.847 & (0.23) & 1.192 & (0.22) & 1.503 & (0.19) \\ 
  7 &          OOB & 0.321 & (0.16) & 0.510 & (0.22) & 0.844 & (0.24) & 1.190 & (0.22) & 1.503 & (0.19) \\ 
   & RFProxIH & NA & NA & NA & NA & NA & NA & NA & NA & NA & NA \\ 
   \hline
Lymphography &      RF-GAP & \textbf{1.046} & (0.14) &\textbf{ 1.521} & (0.13) &\textbf{ 2.480} & (0.13) & \textbf{3.599} & (0.12) & 4.569 & (0.12) \\ 
  148 &    Original & 1.065 & (0.15) & 1.551 & (0.14) & 2.529 & (0.14) & 3.650 & (0.13) & \textbf{4.568} & (0.12) \\ 
  18 &          OOB & 1.065 & (0.14) & 1.549 & (0.13) & 2.524 & (0.13) & 3.644 & (0.13) & \textbf{4.568} & (0.13) \\ 
   & RFProxIH & 1.076 & (0.15) & 1.567 & (0.14) & 2.551 & (0.14) & 3.670 & (0.13) & 4.579 & (0.12) \\ 
   \hline
Optical Digits &      RF-GAP & \textbf{15.774} & (0.17) & \textbf{22.572} & (0.17) & \textbf{37.003} & (0.19) & \textbf{55.369} & (0.21) & \textbf{71.456} & (0.26) \\ 
  3823 &    Original & 17.222 & (0.20) & 24.612 & (0.19) & 40.138 & (0.20) & 58.897 & (0.22) & 73.616 & (0.27) \\ 
  64 &          OOB & 17.199 & (0.19) & 24.549 & (0.18) & 39.904 & (0.20) & 58.569 & (0.21) & 73.442 & (0.27) \\ 
   & RFProxIH & 17.303 & (0.20) & 24.732 & (0.19) & 40.336 & (0.20) & 59.062 & (0.22) & 73.681 & (0.27) \\ 
   \hline
Parkinsons &      RF-GAP & \textbf{0.400} & (0.18) &\textbf{ 0.60}0 & (0.20) &\textbf{ 0.968} & (0.19) & \textbf{1.471} & (0.16) & 1.841 & (0.15) \\ 
  197 &    Original & 0.411 & (0.18) & 0.613 & (0.20) & 0.988 & (0.18) & 1.484 & (0.16) & 1.846 & (0.15) \\ 
  23 &          OOB & 0.408 & (0.18) & 0.610 & (0.20) & 0.979 & (0.19) & 1.475 & (0.16) & \textbf{1.836} & (0.15) \\ 
   & RFProxIH & 0.412 & (0.18) & 0.616 & (0.20) & 0.994 & (0.18) & 1.489 & (0.16) & 1.851 & (0.15) \\ 
   \hline
Seeds &      RF-GAP & \textbf{0.204} & (0.04) & \textbf{0.283} & (0.04) &\textbf{ 0.476} & (0.04) & \textbf{0.703} & (0.04) & \textbf{0.895} & (0.04) \\ 
  210 &    Original & 0.223 & (0.05) & 0.308 & (0.04) & 0.506 & (0.04) & 0.730 & (0.04) & 0.911 & (0.04) \\ 
  7 &          OOB & 0.218 & (0.05) & 0.300 & (0.04) & 0.494 & (0.04) & 0.715 & (0.04) & 0.897 & (0.04) \\ 
   & RFProxIH & 0.226 & (0.05) & 0.313 & (0.04) & 0.512 & (0.04) & 0.737 & (0.04) & 0.917 & (0.04) \\ 
   \hline
Sonar &      RF-GAP & \textbf{3.044 }& (0.13) & \textbf{4.551} & (0.13) &\textbf{ 7.278 }& (0.14) & \textbf{10.879} & (0.16) &\textbf{ 13.923} & (0.18) \\ 
  208 &    Original & 3.246 & (0.13) & 4.834 & (0.14) & 7.631 & (0.15) & 11.174 & (0.16) & 14.055 & (0.18) \\ 
  60 &          OOB & 3.211 & (0.13) & 4.778 & (0.13) & 7.544 & (0.15) & 11.101 & (0.16) & 14.011 & (0.18) \\ 
   & RFProxIH & 3.264 & (0.14) & 4.864 & (0.14) & 7.673 & (0.15) & 11.200 & (0.16) & 14.069 & (0.18) \\ 
   \hline
Titanic &      RF-GAP & \textbf{0.409} & (0.22) & \textbf{0.609} & (0.23) & \textbf{1.067} & (0.26) & \textbf{1.600} & (0.22) & \textbf{1.991} & (0.16) \\ 
  712 &    Original & 0.416 & (0.21) & 0.621 & (0.23) & 1.096 & (0.26) & 1.639 & (0.22) & 2.010 & (0.16) \\ 
  7 &          OOB & 0.415 & (0.21) & 0.620 & (0.23) & 1.093 & (0.26) & 1.637 & (0.22) & 2.010 & (0.16) \\ 
   & RFProxIH & NA & NA & NA & NA & NA & NA & NA & NA & NA & NA \\ 
\hline
Wine &      RF-GAP & \textbf{0.236} & (0.08) & 0.345 & (0.07) & 0.579 & (0.07) & 0.825 & (0.07) & 1.044 & (0.08) \\ 
  178 &    Original & \textbf{0.236} & (0.08) & 0.344 & (0.08) & \textbf{0.577} & (0.07) & \textbf{0.822} & (0.07) & \textbf{1.038} & (0.08) \\ 
  13 &          OOB & 0.240 & (0.08) & 0.353 & (0.07) & 0.585 & (0.07) & 0.829 & (0.07) & 1.043 & (0.08) \\ 
   & RFProxIH & 0.237 & (0.08) & \textbf{0.343} & (0.08) &\textbf{ 0.577} & (0.07) & 0.823 & (0.07) & 1.040 & (0.07) \\ 
   \hline
\end{tabular}

\label{tab:imp}
\end{table}

Here we show extended results on data imputation using random forest proximities. Table~\ref{tab:imp} shows the average imputation results for 100 trials across 17 datasets from Table~\ref{tab:datasets}~\footnote{Larger datasets were not compared here as the PBK and RFProxIH proximities are computationally inefficient. Datasets that were already missing values were also not used.} from the UCI repository~\cite{UCI2019}  using four of the proximity measures with a single iteration. For each dataset, values were removed completely at random in amounts of 5\%, 10\%, 25\%, 50\%, and 75\%. The PBK proximities were omitted from this study due to their slow computational complexity. Additionally, some datasets were not compatible with RFProxIH due to continuous responses or categorical features. 

Missing values were imputed using the three proximity definitions and only a single iteration. The experiment was repeated 100 times for each dataset. For each repetition, the same missing data was used across each of the three proximity types.
The mean squared error (MSE) is almost universally lower when using RF-GAP for imputation.  Even when outperformed, RF-GAP usually ranks second place. Refer to the ranking summary in Table~\ref{tab:imp_ranks}.  The Banknote, Ionosphere, Optical Digits, Sonar, and Waveform datasets particularly show good examples of RF-GAP for imputation. In these examples, RF-GAP outperforms each of the other definitions and the error decreases monotonically.

\begin{figure*}
    \centering
    \includegraphics[width = \textwidth]{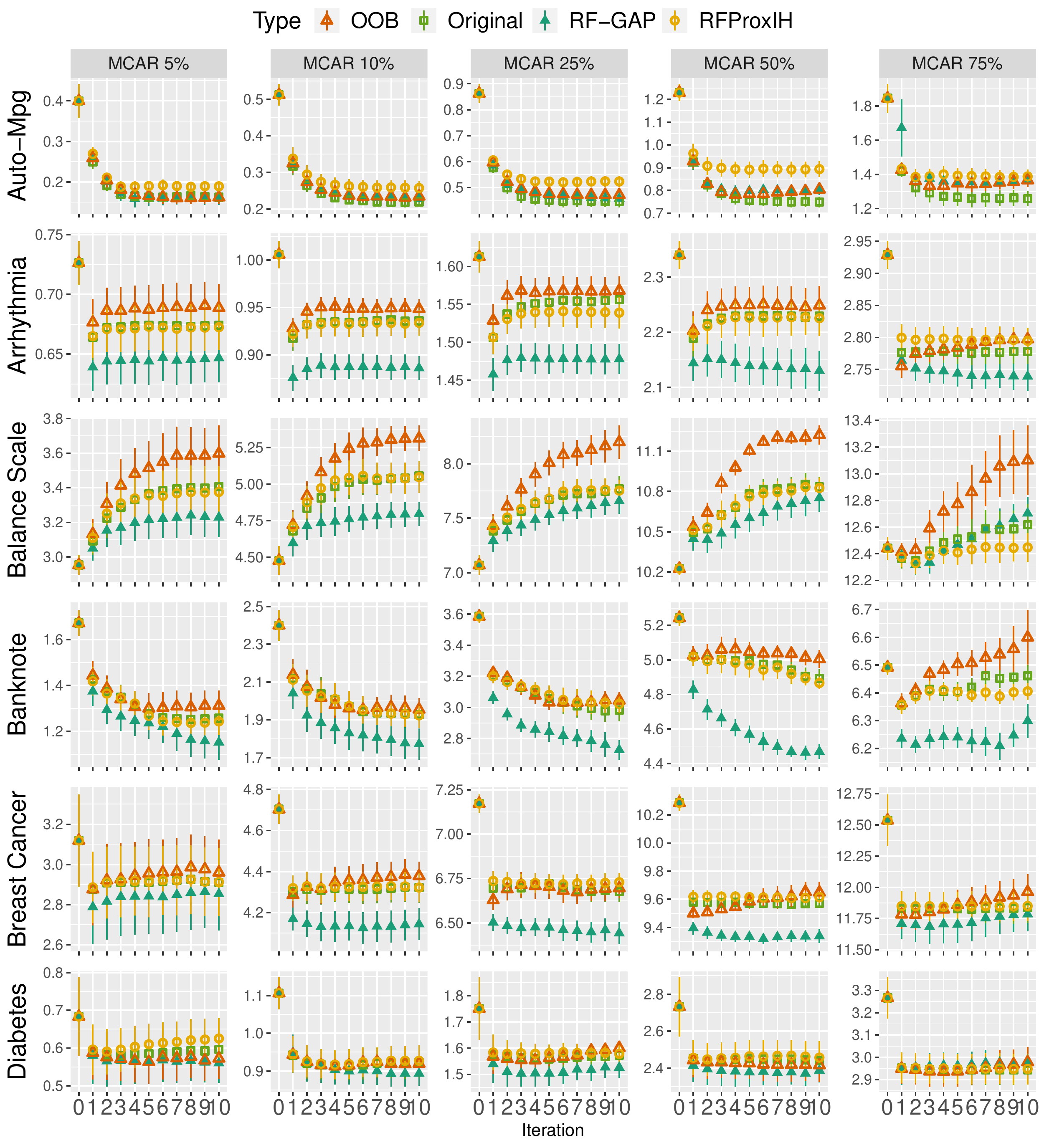}
    \caption{Additional imputation results. See Figure~\ref{fig:imputation_third} for more details.}
    \label{fig:imputation_first}
\end{figure*}

\begin{figure*}
    \centering
    \includegraphics[width = .95\textwidth]{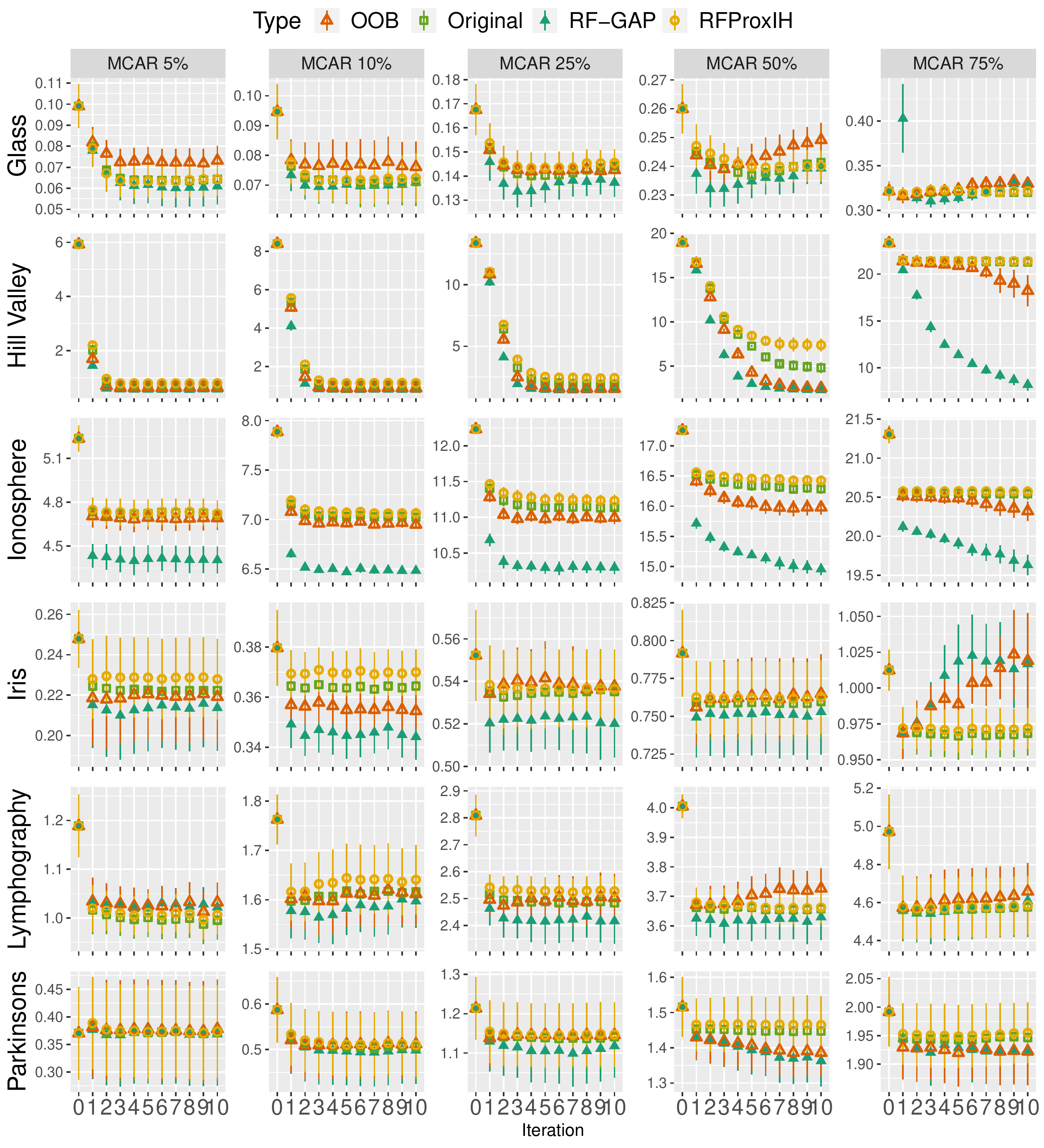}
    \caption{Additional imputation results. See Figure~\ref{fig:imputation_third} for more details.}
    \label{fig:imputation_second}
\end{figure*}

\begin{figure*}
    \centering
    \includegraphics[width = .95\textwidth]{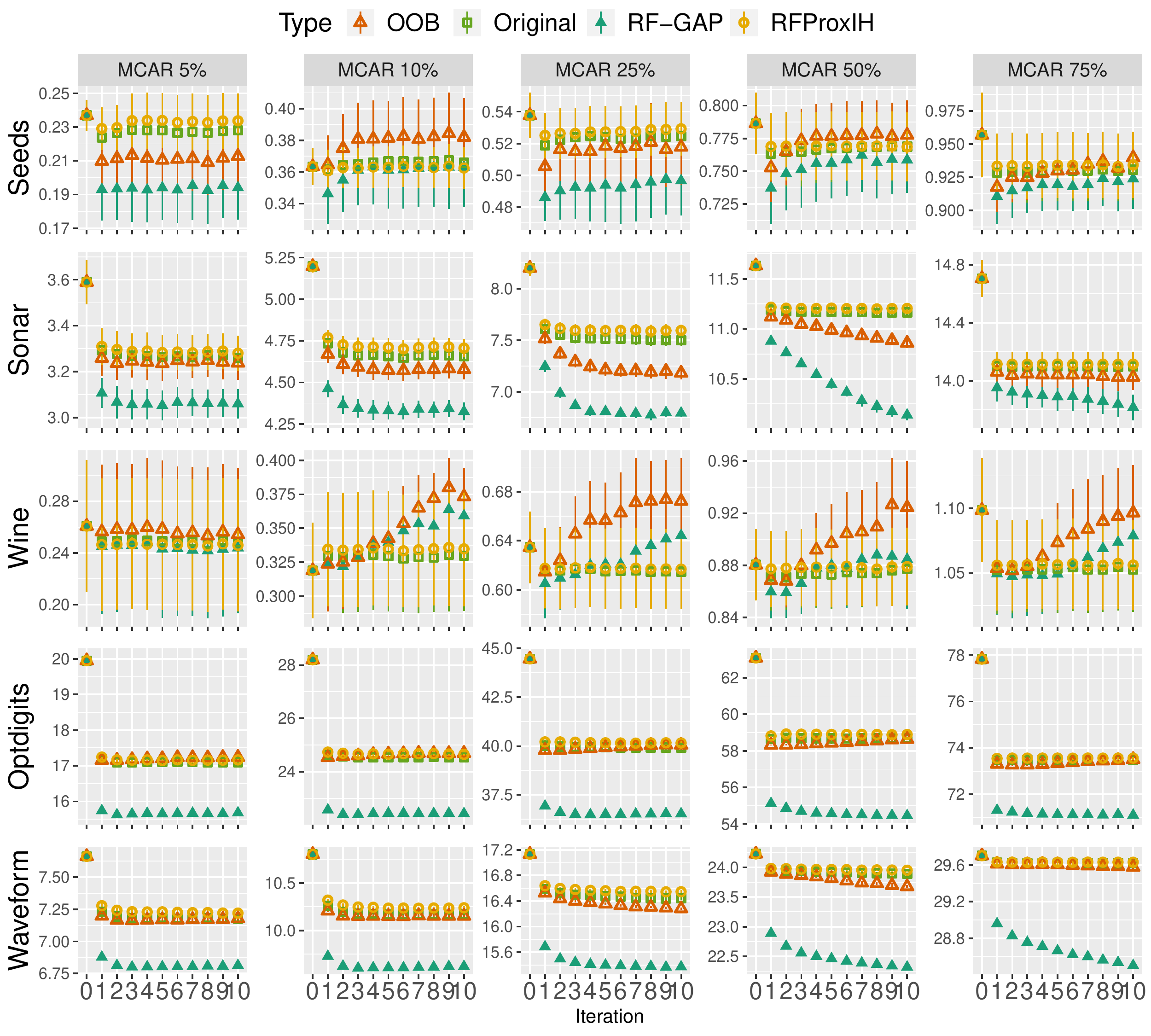}
    \caption{This figure, Figure~\ref{fig:imputation_first} and \ref{fig:imputation_second} give the mean-squared error (MSE) between the original and imputed values using four random forest proximity measures. All data variables were scaled from 0 to 1 for comparison. The scores were compared using 1 to 10 imputation iterations as described in Section \ref{subsec:imputation} and each experiment was conducted over 5 repetitions using the original proximities, OOB proximities, RF-GAP proximities, and RFProxIH (PBK was too inefficient for comparison). Iteration 0 provides the MSE for the median-filled imputation. Five different percentages of values missing completely at random (MCAR) were used (5\%, 10\%, 25\%,  50\%, and 75\%) across several datasets found in Table~\ref{tab:datasets}. In general, RF-GAP usually outperforms the other proximity-weighted imputations. See additional results in Table~\ref{tab:imp}.}
    \label{fig:imputation_third}
\end{figure*}

Figures \ref{fig:imputation_first}, \ref{fig:imputation_second}, and \ref{fig:imputation_third} show imputation results across multiple iterations. Each experiment was repeated 5 times across 10 iterations. In general, RF-GAP outperforms the other proximity-weighted imputations although the imputation tends to be much noisier for smaller datasets (see Balance Scale, Iris, Seeds, and Wine, for example) and less reliable for large percentages of missing values.  This is particularly prominent when 75\% of the data is missing. 

\newpage

\section{Dataset Descriptions}

The majority of datasets used in this paper come directly from the UCI repository. See a full list of dataset descriptions in Table~\ref{tab:datasets}.

\begin{table}[!htb]
    \centering
    \caption{Descriptions of the datasets used in the experiments for this paper. The number of observations accounts for the removal of any observations which had missing values.  The number of columns accounts for the removal of any uniquely-identifying variables (i.e., variables which are distinct for each observation).}
    \begin{tabular}{|c|c|c|c|c|c|}
    \hline
        Data & Label Type & No. Obs. & No Vars. & Missing Values? & Source  \\
    \hline
        Arrhythmia & Continuous & 452 & 279 & No  & UCI  \\
    \hline
        Auto-MPG & Continuous & 392 & 8 & Yes & UCI  \\
    \hline
        Balance Scale  & Categorical & 645 & 4 & No & UCI  \\
    \hline
        Banknote  & Categorical & 1372 & 5 & No & UCI \\
    \hline
        Breast Cancer  & Categorical & 699 & 16 & Yes & UCI \\
    \hline
        Car  & Categorical & 1728 & 6 & No & UCI \\
    \hline
        Diabetes  & Categorical & 678 & 8 & No & UCI \\
    \hline
        E. Coli  & Categorical & 336 & 8 & No & UCI  \\
    \hline
        Glass  & Categorical & 214 & 10 & No & UCI  \\
    \hline
        Heart Disease  & Categorical & 303 & 13 & Yes & UCI \\
    \hline
        Hill Valley  & Categorical & 606 & 101 & No & UCI  \\
    \hline
        Ionosphere  & Categorical & 351 & 34 & No & UCI  \\
    \hline
        Iris  & Categorical & 150 & 4 & No & UCI  \\
    \hline
        Liver  & Categorical & 345 & 7 & No & UCI \\
    \hline
        Lymphography  & Categorical & 148 & 18 & No & UCI  \\
    \hline
        Optical Digits  & Categorical & 3823 & 64 & No & UCI  \\
    \hline
        Parkinson's  & Categorical & 197 & 23 & No & UCI  \\
    \hline
        RNA-Seq (Gene Expression)  & Categorical & 801 & 20531 & No & UCI  \\
    \hline
        Seeds   & Categorical & 210 & 7 & No & UCI  \\
    \hline
        Sonar (Connectionist Bench)  & Categorical & 208 & 60 & No & UCI  \\
    \hline
        Tic-Tac-Toe  & Categorical & 958 & 9 & No & UCI  \\
    \hline
        Titanic  & Categorical & 712 & 7 & Yes & Kaggle  \\
    \hline
        Waveform  & Categorical & 5000 & 21 & No & UCI  \\
    \hline
        Wine  & Categorical & 178 & 13 & No & UCI \\
    \hline
    \end{tabular}

    \label{tab:datasets}
\end{table}

\end{document}